\documentclass[11pt, a4paper, logo, twocolumn, copyright]{googledeepmind}

\usepackage[authoryear, sort&compress, round]{natbib}
\usepackage{longtable}
\usepackage[none]{hyphenat}
\usepackage{xurl}
\usepackage{hyperref}
\usepackage{graphicx}
\graphicspath{ {./figures/} }
\numberwithin{figure}{section}
\title{Sociotechnical Safety Evaluation of Generative AI Systems}

\correspondingauthor{Laura Weidinger <lweidinger@deepmind.com>}

\paperurl{arxiv.org/abs/123}

\reportnumber{001} % Leave blank if n/a

\renewcommand{\today}{2000-01-01}

\author[1]{Laura Weidinger}
\author[1]{Maribeth Rauh}
\author[1]{Nahema Marchal}
\author[1]{Arianna Manzini}
\author[1]{Lisa Anne Hendricks}
\author[1]{Juan Mateos-Garcia}
\author[1]{Stevie Bergman}
\author[1]{Jackie Kay}
\author[1]{Conor Griffin}
\author[1]{Ben Bariach}
\author[1]{Iason Gabriel}
\author[1]{Verena Rieser}
\author[1]{William Isaac}

\affil[1]{Google DeepMind, London N1C 4DN, United Kingdom}

\begin{abstract} \label{abstract}

Generative AI systems produce a range of risks. To ensure the safety of generative AI systems, these risks must be evaluated. In this paper, we make two main contributions toward establishing such evaluations. First, we propose a three-layered framework that takes a structured, sociotechnical approach to evaluating these risks. This framework encompasses capability evaluations, which are the main current approach to safety evaluation. It then reaches further by building on system safety principles, particularly the insight that context determines whether a given capability may cause harm. To account for relevant context, our framework adds human interaction and systemic impacts as additional layers of evaluation. Second, we survey the current state of safety evaluation of generative AI systems and create a repository of existing evaluations. Three salient evaluation gaps emerge from this analysis. We propose ways forward to closing these gaps, outlining practical steps as well as roles and responsibilities for different actors. Sociotechnical safety evaluation is a tractable approach to the robust and comprehensive safety evaluation of generative AI systems.

\end{abstract}

\begin{document}

\maketitle

\onecolumn

\section*{Acknowledgements}

We thank Simon Osindero, Sasha Brown, Matt Botvinick, Canfer Akbulut, Suresh Venkatasubramanian, Victor Ojewale, Fernando Diaz, Olivia Wiles, Doug Fritz, Courtney Biles, Nicklas Lundblad, Neil Rabinowitz, Jenny Brennan, Sunipa Dev, Don Wallace, Ramona Comanescu, Mark Díaz, Michal Lahav, Alex Kaskasoli, Isabela Albuquerque, Seliem El-Sayed, and Rida Qadri for their feedback and contributions to this work.

\onecolumn

\tableofcontents

\clearpage

\subsection*{Reader’s guide}

\begin{figure}[h]
    \centering
    \caption{A sociotechnical framework for safety evaluation comprises three layers.}
    \label{fig1}
    \includegraphics[width=6cm]{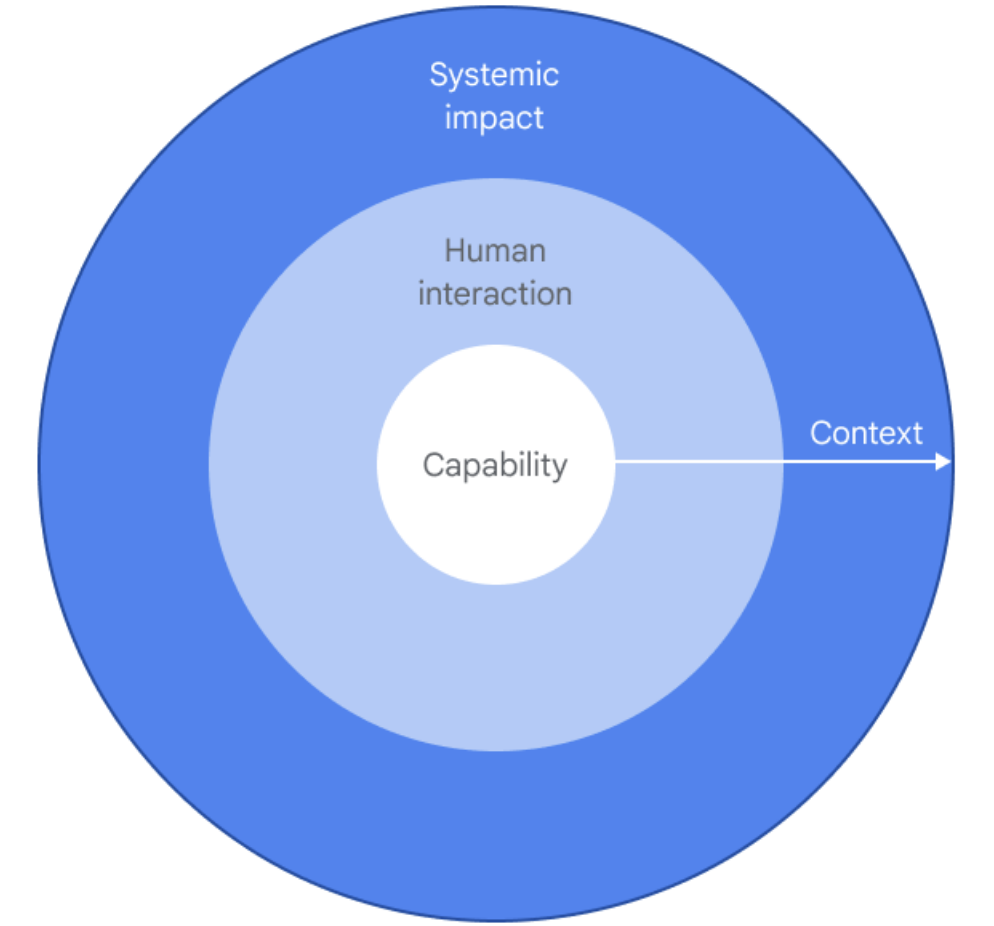}
\end{figure}

\begin{figure}[h]
    \centering
    \caption{Evaluations per harm area and AI system output modality. No harm area is well covered across modalities.}
    \label{fig2}
    \includegraphics[width=8cm]{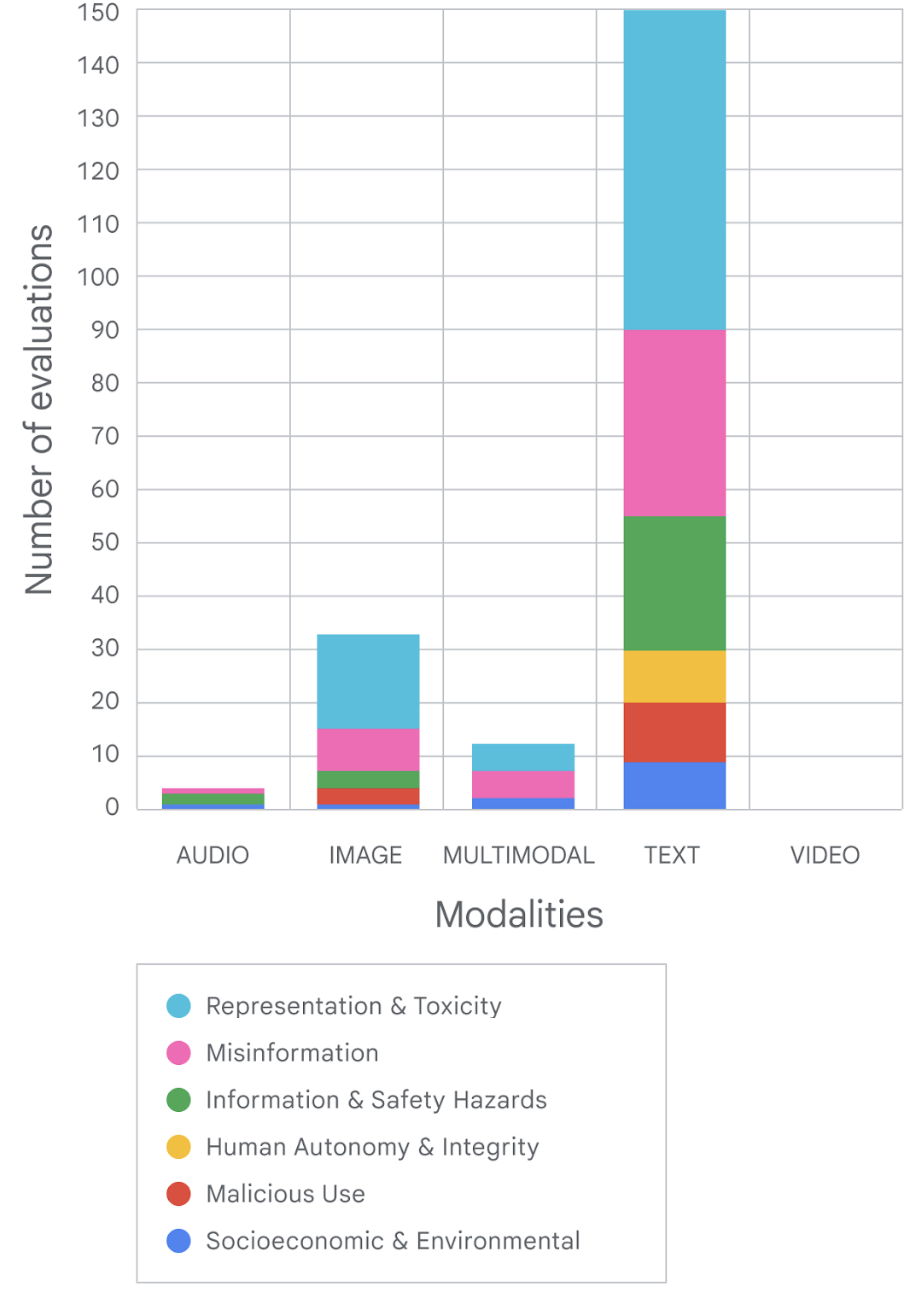}
\end{figure}

This is a long document. Depending on your background and interests, we recommend different reading strategies:

\begin{itemize}
  \setlength\itemsep{1em}

  \item \textbf{Two-minute read:} Look at \hyperref[fig1]{figure 2.1} (p.\pageref{fig1}) that illustrates our three-layered evaluation framework, and \hyperref[fig2]{figures 3.1-3} (p.\pageref{fig2}) which depict the current state of safety evaluations.

  \item \textbf{Ten-minute read:} Read the abstract and skim \hyperref[section2]{section 2} (p.\pageref{section2}), which introduces our three-layered evaluation framework; look at \hyperref[fig2]{figures 3.1-3} (p.\pageref{fig2}) which depict the current state of safety evaluations.

  \item \textbf{Evaluators:} Skim \hyperref[section2]{section 2} (p.\pageref{section2}), where we introduce our three-layered evaluation framework, and \hyperref[section3]{section 3} (p.\pageref{section3}) where we survey the current state of safety evaluation; dedicate most time to \hyperref[section4]{section 4} (p.\pageref{section4}) on practical steps to closing evaluation gaps, and to the \hyperref[casestudy1]{case study} (p.\pageref{casestudy1}) on evaluating misinformation that puts our evaluation framework into practice. Read about evaluation as a practice of responsible innovation in the \hyperref[discussion]{discussion} (p.\pageref{discussion}), and about the limitations of specific \hyperref[appendixa2]{evaluation methodologies} (p.\pageref{appendixa2}).

  \item \textbf{People steering AI labs:} Look at \hyperref[fig1]{figure 2.1} (p.\pageref{fig1}) that illustrates our three-layered evaluation framework, read \hyperref[section3]{section 3} (p.\pageref{section3}), which outlines gaps in the current state of safety evaluation of generative AI systems, and look at \hyperref[fig5]{figure 5} (p.\pageref{fig5}) that illustrates the roles \& responsibilities of different actors. Consider the limitations of evaluation methods laid out in the section on \hyperref[appendixa2]{evaluation methodologies} (p.\pageref{appendixa2}) and our \hyperref[casestudy1]{case study} (p.\pageref{casestudy1}) on evaluating misinformation that puts our evaluation framework into practice.

  \item \textbf{Public policy makers:} Look at \hyperref[fig1]{figure 2.1} (p.\pageref{fig1}) that illustrates our three-layered evaluation framework, skim \hyperref[section3]{section 3} (p.\pageref{section3}), which lays out the state of evaluation today; and read the part on \hyperref[rolesandresponsibilities]{roles and responsibilities} (p.\pageref{rolesandresponsibilities}) in the  \hyperref[discussion]{discussion section}.  

  \item \textbf{AI researchers:} Consider the evaluation framework in \hyperref[section2]{section 2} (p.\pageref{section2}), concrete ways forward as introduced in \hyperref[section4]{section 4} and in the \hyperref[casestudy1]{case study} (p.\pageref{casestudy1}), and the limitations and implications in the \hyperref[discussion]{discussion section} (p.\pageref{discussion}).

\end{itemize}

\twocolumn

\section{Introduction}

Generative\footnote{By “generative” we refer to AI systems that generate novel output rather than analysing existing data \citep{gpt-3_generative_2022}. We focus on generative AI systems, which we define as models that input and output any combination of image, audio, video, and text. This includes transformer-based systems, such as large language models, diffusion-based systems, and hybrid architectures.}, multimodal\footnote{By “multimodal” we refer to models that accept and produce output in any combination of image, audio, and text. This includes models that accept or produce output in more than one modality, such as interleaved image and text data, or audiovisual data.} AI systems\footnote{By “AI system” we refer to a pre-trained base model or foundation model, potentially “fine-tuned” by adapting it to particular datasets for specific performance targets, including via practices such as RLHF. AI systems may also include filters such as input or output filters. AI systems are ready for integration into a product.} are becoming increasingly widely used. Real-world applications of generative AI systems are proliferating across domains, ranging from medical applications \citep{singhal_towards_2023, nori_capabilities_2023} to news and politics (e.g. \citet{bruell_buzzfeed_2023}) and social interaction such as companionship (e.g. \citet{pentina_exploring_2023, griffith_my_2023}). Early systems produced output in single modalities, such as image generation \citep{rombach_high-resolution_2022, ramesh_zero-shot_2021} and text capabilities, producing compelling natural language \citep{openai_gpt-4_2023-1, glaese_improving_2022, anil_palm_2023}. Increasingly, generative AI systems in other modalities such as audio, including voice and music \citep{oord_wavenet:_2016, dhariwal_jukebox:_2020, borsos_audiolm:_2023, agostinelli_musiclm:_2023, huang_noise2music:_2023}, and video and audiovisual capabilities are steadily improving \citep{du_conditional_2023}. Generative AI systems are increasingly multimodal, and their integration into various aspects of life is anticipated \citep{noauthor_google_2023}.

In addition to creating benefits, generative AI systems pose risks of harm. For individual modalities, these risks have been mapped out in different taxonomies \citep{bommasani_opportunities_2022, weidinger_ethical_2021, liu_trustworthy_2023, bird_typology_2023, solaiman_evaluating_2023, shevlane_model_2023, barnett_ethical_2023, shelby_sociotechnical_2023, dinan_anticipating_2021} as well as in research on individual risks or applications (e.g. \citet{luccioni_stable_2023, bianchi_easily_2023, birhane_multimodal_2021, carlini_extracting_2023, shevlane_model_2023, khlaaf_hazard_2022}). Complementing foresight research, observed instances of harm from generative AI systems have been logged to identify risks that these systems create \citep{noauthor_welcome_nodate, noauthor_expert_nodate}. Now that risks from generative AI systems have been identified, their impact on the overall safety of a generative AI system must be understood. This requires evaluation.

The growing use of generative AI systems makes it both easier and more pressing to evaluate potential risks of harm. As these technologies become widely used and embedded, the risks they create are a public safety concern. Accordingly, evaluating potential risks from generative AI systems is a growing priority for AI developers \citep{openai_our_nodate, noauthor_core_2023}, public policy makers \citep{the_white_house_fact_2023}, regulators \citep{noauthor_initial_nodate, noauthor_regulatory_2023, noauthor_ai_2021, national_institute_of_standards_and_technology_ai_nodate}, and civil society \citep{noauthor_regarding_nodate}. 

Evaluation is the practice of measuring AI system performance or impact. Safety evaluation in particular focuses on evaluating risks of harm or actualised impacts on people or broader systems. Evaluations can be exploratory (such as open-ended probing of an AI system) or directed (such as running a specific test). They include qualitative investigations, such as studying how people actually attempt to use an AI system, as opposed to assessing intended use cases. Exploratory evaluations may identify areas of uncertainty or additional context, or give rise to novel directed evaluation questions. Directed evaluations follow a series of steps, whereby a target – such as a risk of harm – is selected, operationalised into an observable metric, and measured. In any evaluation, the results are then judged against a normative baseline, such as whether an AI system is “good”, “fair”, or “safe enough”. Evaluation is never neutral: it rests on interwoven technical and normative decisions, such as deciding what to evaluate in the first place, how to measure it (see \hyperref[operationalisingrisks]{Operationalising risks}), and what results indicate “good” AI system performance \citep{bakalar_fairness_2021, bowker_sorting_2000}. Safety evaluation can form part of broader safety audits, which may additionally take into account organisational governance structures or existing documentation and more \citep{raji_closing_2020, mokander_auditing_2023, costanza-chock_who_2022}.

Evaluation performs an important function by providing public safety assurances. By systematically testing AI systems against potential risks of harm, evaluation can make AI systems less opaque. Evaluation also sheds light on, predicts, and quantifies the likelihood of potential downstream harms, and can surface the factors and mechanisms that influence whether downstream harm may occur. Evaluations can guide the development of AI systems, as well as providing assurances on levels of AI system safety in different contexts. As a result, the understanding of AI systems that evaluations provide is essential for well-informed, responsible decision-making on AI system development and deployment \citep{stilgoe_developing_2013}. Further, evaluation of different risk areas brings to light normative trade-offs that arise as AI systems are developed and deployed in real-world settings. By performing these functions, evaluation is a foundation for meaningful accountability on the responsible innovation and deployment of generative AI systems. 

In this paper, we make two main contributions: a sociotechnical framework for safety evaluation and an empirical assessment of the current safety evaluation landscape. While the priority of safety evaluations for generative AI systems is clear, current approaches are often heterogeneous and ad hoc. The evaluations that are being conducted differ between organisations and AI systems (e.g. \citet{glaese_improving_2022,anthropic_model_nodate, openai_gpt-4_2023, mishkin_dalle_2022, anil_palm_2023}), which makes them hard to compare and reproduce. This can also mean that the evaluation of a given AI system misses important risks that should be considered. We argue that a more systematic and standardised approach to safety evaluation is necessary to ensure meaningful, comparable, and comprehensive safety evaluation (c.f. \citet{naiac2023, liang_holistic_2022, noauthor_u.s._2019}). As a step in that direction, we offer a sociotechnical framework to guide safety evaluation of generative AI systems. 

Our second main contribution is a review of the current state of safety evaluations for generative AI systems, including the public release of a \href{https://dpmd.ai/46CPd58}{repository} of existing evaluations, and analysing gaps. These gaps are tractable: we present practical steps toward closing them. We propose roles and responsibilities for different stakeholders and discuss how currently disparate communities with interests in the safety of AI systems can intersect.

The paper proceeds as follows. In \hyperref[section2]{section 2}, we outline our proposed framework for sociotechnical safety evaluation across three layers that progressively add context: the capability layer, the human interaction layer, and the systemic impact layer. \hyperref[section3]{Section 3} surveys the current state of research and practice in sociotechnical safety evaluation, identifying strengths and limitations of existing approaches. \hyperref[section4]{Section 4} builds on the framework and survey to discuss ways to close observed evaluation gaps. In \hyperref[discussion]{section 5}, we conclude with a discussion of open questions for the field of safety evaluations, including the varying roles and responsibilities between AI developers and public sector stakeholders for conducting evaluations across the layers, and the connections between the range of proposed approaches to safety associated with generative AI systems.

\section{Framework for sociotechnical AI safety evaluation} \label{section2}

Recent research identified a sociotechnical gap in our understanding of the safe development and deployment of AI systems \citep{lazar_ai_2023, shelby_sociotechnical_2023, mohamed_decolonial_2020, selbst_fairness_2019}. This sociotechnical gap arises where AI system safety is evaluated only with regard to technical components of an AI system, i.e. individual technical artefacts such as data, model architecture, and sampling strategies. While these are important aspects of AI safety evaluation, they alone are insufficient to determine whether an AI system is safe. Instead, an approach is needed that takes into account human and systemic factors that co-determine risks of harm. 

To close this gap, we apply a sociotechnical lens to AI safety evaluation. Sociotechnical research has a long-standing history in expanding the frontiers of AI system evaluation to include human and systemic factors \citep{raji_closing_2020, ekstrand2018all, dwork2012fairness, friedman1996bias, barocas2016big}.  This approach is rooted in the observation that AI systems are sociotechnical systems: both humans and machines are necessary in order to make the technology work as intended \citep{selbst_fairness_2019}. The interaction of technical and social components determines whether risk manifests \citep{leveson_engineering_2012}. Consequently, AI evaluation requires a framework that integrates these components and their interactions. 

Similarly to other sociotechnical work (c.f. \cite{raji_closing_2020, rismani2023beyond}), our approach is further inspired by a system safety approach from the discipline of safety engineering. System safety represents  a historical paradigm shift in safety engineering, from component-based approaches toward systems thinking, taking into account broader contexts, interactions and emergent properties of complex systems \citep{leveson_engineering_2012}. Component-based approaches to safety emerged historically from industries dealing with hazardous materials in constrained settings and processes. These approaches isolate system components or steps in a process and assess individual failure modes of each part. The sum of these assessments is then considered a comprehensive safety evaluation of the entire system. This approach is not fit for purpose in complex and versatile systems, such as where software and people interact with great degrees of freedom \citep{hutchins_how_1995, leveson_engineering_2012}. Here, a component-based approach to safety needs to give way to a system-based approach to safety \citep{leveson_engineering_2012}. An analogous shift is required in the safety evaluation of generative AI systems. We propose a framework that accounts for this shift.

Specifically, we present a three-layered framework to structure safety evaluations of AI systems. The three layers are distinguished by the target of analysis. The layers are: \textit{capability} evaluation, \textit{human interaction} evaluation, and \textit{systemic impact} evaluation. These three layers progressively add on further context that is critical for assessing whether a capability relates to an actual risk of harm or not. 

To illustrate these three evaluation layers, consider the example of misinformation harms. \textit{Capability} evaluation can indicate whether an AI system is likely to produce factually incorrect output (e.g. \citet{lin_truthfulqa:_2022, ji_survey_2023}). However, the risk of people being deceived or misled by that output may depend on factors such as the context in which an AI system is used, who uses it, and features of an application (e.g. whether synthetic content is effectively signposted). This requires evaluating \textit{human-AI interaction}. Misinformation risks raise concerns about large-scale effects, for example on public knowledge ecosystems or trust in shared media. Whether such effects manifest depends on \textit{systemic} factors, such as expectations and norms in public information sharing and the existence of institutions that provide authoritative information or fact-checking. Evaluating misinformation risks from generative AI systems requires evaluation at each of these three layers.

The three layers in this framework interact and their boundaries are gradual. Effects detected at one layer may be indicate related observations at the next. For example, discrepancies in how an AI system performs for different user groups can be identified at the layer of human interaction, and may foreshadow disparate systemic impacts for these groups. A further illustration of the gradual boundaries between the layers is that evaluation methods can straddle multiple layers. For example, adversarial testing is a method for evaluating capabilities. However, by focusing on the experience of the adversarial tester, it can be a measure of human interaction: specifically of the friction a person encounters when trying to use an AI system to malicious ends. While interactions between these layers may extend beyond the failure of individual system components and be complex, they are often still within the control of AI system developers. 

In addition, there are feedback loops within and between layers. For example, societal context may feed back into system capabilities via the opinions and demographics of human annotators, as annotated data is used to adapt AI systems to particular contexts. Note that the layers are not ordered by importance nor in any chronological order. Rather, evaluations at each layer can be performed simultaneously and asynchronously. We now introduce the three-layered framework in detail.

\subsection{Layer 1: Capability}

Evaluation at this layer targets AI systems and their technical components.\footnote{These include training data; model components, such as model architectures and classifiers; the model itself, such as pre-training embeddings; and model outputs, such as images.} These are routinely evaluated in isolation, including tests of AI system behaviour in response to novel tasks or environments, or testing individual technical artefacts, such as the data that an AI system is trained on. It also includes evaluating processes by which these artefacts are created, such as the aggregation mechanisms in processes that are used to adapt an AI system to a particular task. In addition to assuring the safety of an AI system, evaluations at this layer are often performed to guide iterative model development ("hill-climbing").

While evaluation at this layer does not assess downstream harm per se, it can provide indication of whether a component, output, or AI system is likely to cause downstream harm. Several risks of harm can be evaluated by measuring capabilities through the outputs of an AI system. This includes, for example, the extent to which an AI model reproduces harmful stereotypes in images or utterances (representation harms \citep{bianchi_easily_2023}), makes factual errors, or displays advanced capabilities that present safety hazards. The extent to which model performance deteriorates when prompted in different languages, about different groups, or in different domains can also be evaluated at this layer and can be indicative of the likely distribution of potential downstream harm. Capabilities also include metrics that are designed to track efficiency and may shed light on potential downstream environmental impact, such as energy use at inference \citep{kaack_aligning_2022}. Capabilities can be assessed against fixed, automated tests or probed dynamically by human or automated adversarial testers (see \hyperref[selectevalmethods]{Selecting evaluation methods}).

Evaluations at this layer can also concern the data on which a model is trained. Using tools to visualise clusters or associations in the training data can reveal diversity and representativeness of the data, or the presence of sensitive data such as private information \citep{dodge_documenting_2021, choi_tools_2023, wang_measuring_2022, kreutzer_quality_2022}. Similar tools can be used to assess the learned associations of a trained AI system \citep{caliskan_semantics_2017, steed_image_2021}. 

Other components that can be analysed at this layer include filters and techniques used to reduce output that may relate to a particular risk of harm, such as filters for safety harms in images \citep{rando_red-teaming_2022} or toxic language \citep{noauthor_perspective_nodate}. However, such filters have limitations \citep{rando_red-teaming_2022} that can aggravate representation harm by disproportionately filtering out content from some groups \citep{welbl_challenges_2021}. 

Capability evaluation is critical, but insufficient, for a comprehensive safety evaluation. It can serve as an early indicator of potential downstream harms, but to assess whether or not a capability relates to risks of harm requires taking into account context – such as who uses the AI system, to what end, and under which circumstances. This context is assessed at subsequent layers.

\subsection{Layer 2: Human interaction}

This layer centres the experience of people interacting with a given AI system. Assessing AI system safety requires evaluating not just the AI system in isolation but also effects on people interacting with AI systems, and the human–AI dyad. This includes usability: does the AI system perform its intended function at the point of use, and how do experiences differ between user groups? This layer also centres potential externalities: does human–AI interaction lead to unintended effects on the person interacting or exposed to AI outputs? Evaluation at this layer acknowledges that AI system safety depends on who uses an AI system, with what goal in mind, and in what context. This layer shifts the lens to the humans interacting with an AI system and is key to a human-centred approach to AI development \citep{liao_ai_2023, tahaei_human-centered_2023, vaughan_human-centered_2021}. 

In addition to testing AI system capabilities, the functionality of an AI system in the context of a concrete application must be assessed \citep{raji_fallacy_2022}. This includes testing how different people actually use the system, as real-world use often deviates from intended use cases. User groups are heterogeneous, and safety evaluation requires not only assessing whether an AI system works but also for whom it works well \citep{wang_human-centered_2023}. To assess usability in practice, human interaction with AI systems needs to be evaluated “in the wild”, i.e. in a real-world application context for an AI system such as a hospital \citep{sendak_real-world_2020} or police units \citep{marda_data_2020}. Evaluating the human–AI interaction can also reveal how easy it is to use a model for malicious ends \citep{roy_measuring_2023}. 

Human-centred testing can shed light on potential externalities created by specific use cases or applications of AI systems. To assess a risk of harm, directed psychology or human–computer interaction experiments can be performed. Under controlled, safe conditions, potential harmful outcomes to people interacting with AI can be studied, such as overreliance on AI systems \citep{chiesurin_dangers_2023} or overtrust (e.g. due to AI systems endowed with anthropomorphic cues, \citet{glikson_human_2020}). Some effects may only manifest over time and require longitudinal evaluation. For example, one experiment found that repeated exposure to AI content increases its persuasiveness but only up until a certain point \citep{cacioppo_persuasiveness_1980}. Regarding another risk area, it has also been hypothesised that increased feelings of social isolation due to overuse of technology may only show up after frequent exposure to an AI system, in between a user’s interactions \citep{turkle_alone_2011}. Evaluation at this layer can also assess harms to data annotators, as they are exposed to harmful model outputs, including via surveys or interviews \citep{stoev_experiencing_2023, gray_ghost_2019}. In addition, human interaction evaluations may reveal disparate harm profiles for different modalities. For example, one set of evaluations found that users more readily believe synthetic misinformation that is presented in video as opposed to text modalities \citep{sundar_seeing_2021}. 

Evaluations at this layer can also identify psychological mechanisms by which harms may occur to a person interacting with an AI system. For example, they may identify cognitive biases that influence people coming to believe misinformation \citep{jerit_political_2020}, or how AI systems influence or persuade humans over the course of an interaction, such as when co-writing a text \citep{jakesch_co-writing_2023, hohenstein_artificial_2023}. 

Finally, evaluation that considers an AI system in the context of use can assess the overall performance of the human–AI dyad, such as quality of outcomes on AI-assisted computer coding tasks compared to a human–human baseline \citep{vasconcelos_generation_2023}. 

While this layer provides critical context by adding human interaction to the evaluation, it remains insufficient for a comprehensive AI safety assessment. It provides limited insights on the potential broader impacts that an AI system may have when deployed at scale, and does not consider risks and impacts on broader systems such as society, economic impacts, or the natural environment. Assessing these effects requires analysing the broader systems into which an AI system is deployed, at the third and final layer of our sociotechnical framework for safety evaluation.

\subsection{Layer 3: Systemic impact}

The third target of evaluation is the impact of an AI system on the broader systems in which it is embedded, such as society, the economy, and the natural environment. Widely used AI systems shape, and are shaped by, the societies in which they are used \citep{matias_humans_2023, wagner_measuring_2021}. Detecting the effects from these interactions requires evaluation at the system layer. Some effects may only emerge as an AI system is deployed at large scale. For example, risks from increasing homogeneity in knowledge production and creativity due to “algorithmic monocultures” are emergent at the systems layer of evaluation \citep{toups_ecosystem-level_2023, kleinberg_algorithmic_2021, doshi_generative_2023}. Harms may also have small effect sizes that are hard to detect at the layer of individual user interactions but become salient at a systems level (e.g. \citet{noauthor_social_nodate}). Evaluating these risks and impacts requires focusing on the broader systems into which the AI system is integrated. 

Evaluation at this layer can target systems of different domains and sizes. Economic assessments may concern broad systemic impacts, such as the labour market impacts of generative AI \citep{tolan_measuring_2021, frank_toward_2019, frey_future_2013, felten_occupational_2021, eloundou_gpts_2023} or the impact of model adoption on productivity \citep{brynjolfsson_generative_2023}. It may also centre specific industries or goods, for example by evaluating impacts on the creative economy or predicting the likely impacts of generative AI on the erosion of public goods such as the creative commons \citep{del_rio-chanona_are_2023, huang_generative_nodate}. Impact from generative AI systems on societal institutions, such as political polarisation or changes to trust in public media, can be evaluated through system evaluation \citep{lorenz-spreen_systematic_2023}. The fairness of how benefits and risks are distributed can also be ascertained at this layer, for example by assessing take-up of AI tools across countries \citep{calvino_portrait_2023} and identifying who is able to capture and extract most value using these technologies \citep{brynjolfsson_generative_2023}. 

Evaluations at this layer may also focus on smaller, more localised systems, such as assessing impacts from an AI system in a clinical context on the provision of care \citep{elish_repairing_2020}. Evaluating how AI systems are socially embedded can shed light on how people come to trust the outputs – for example, where friends and colleagues all use a system, this system may be trusted more. One common concern with the widespread availability of generative AI systems is that they can be used to cheat on school assignments \citep{rudolph_chatgpt:_2023}. System-level evaluation of adoption and perception of AI systems can evaluate what types of use occur and under what circumstances, and by whom they constitute 'cheating'. Environmental impacts can be targeted at this layer, to provide a nuanced understanding of impacts on broader ecosystems. For example, detailed and localised evaluation can reveal the actual environmental impact from generative AI systems, such as from data centres that rely on nearby water sources for cooling \citep{sasha_luccioni_estimating_2022}. Early stage indicators, such as energy use as a proxy for environmental impact at the capability layer, can be calibrated and contextualised via evaluations at this layer. 

Evaluation at this layer can also provide context-rich assessments of interactions of different systems as social, economic, and ecological factors intersect. For example, evaluation at the system layer may take into account the biodiversity and resilience of local ecosystems, the nature of the energy grid, and the social and economic implications for local communities in order to assess overall harm of infrastructure that powers an AI system \citep{solaiman_evaluating_2023}.

Systemic impacts are often difficult to assess due to the complex nature, idiosyncrasies, and noise of the systems that are being evaluated. While direct impacts of an AI system may not be known until post deployment, forecasts or comparable technologies can provide initial insights on potential risks of harm at this layer.

\subsection{Summary}

We present a three-layered sociotechnical framework for safety evaluation of generative AI systems. (While we focus on generative AI systems, this framework may also be applicable to other types of AI.) The same high-level risk areas can be detected and evaluated at each layer (we outline practical steps toward evaluating risks at each layer in \hyperref[section4]{section 4}). What connects the three layers is that they progressively add on further context. Note that they are not sequential or dependent on each other. Neither are the layers conditional on each other; rather, evaluation at each layer can be run simultaneously\label{section5}. Integrating results from all layers provides a comprehensive evaluation of the safety of a generative AI system. The layers are a guiding structure to facilitate evaluation along different layers of context in a sociotechnical system.

\section{Current state of sociotechnical safety evaluation} \label{section3}

In this section, we survey the state of safety evaluation of generative AI systems. This first requires consolidating a taxonomy of potential harm from these AI systems. We present a synthesised taxonomy of harm based on prior literature of taxonomies of harm from generative AI systems. Next, we employ an extensive process to identify all existing evaluations of generative AI systems that speak to risks identified in our taxonomy. We map all identified evaluations by risk area; by AI system modalities (image, audio, video, text, and multimodal combinations); and by layers of evaluation based on the three-layered framework introduced above. This mapping is presented in an overview figure that snapshots the sociotechnical evaluation landscape today. We close this section by discussing the “evaluation gaps” this mapping has surfaced.

\subsection{Taxonomy of harm}

First, to assess the state of sociotechnical safety evaluation for generative AI systems requires grounding the types of risk that such evaluation should assess. To this end, we revisit the growing literature on social, ethical, and other safety risks from generative AI systems and integrate insights from this literature into a single, holistic taxonomy (high-level version, \hyperref[tab:table1]{table 1}; detailed version, \hyperref[tab:table2]{appendix section A.1}). Rather than presenting a novel research artefact, the goal of this taxonomy is to provide a basis for mapping the state of safety evaluation of generative AI systems. 

Previous work identified a wide range of risks posed by generative AI systems. Existing taxonomies address risks from AI systems audio \citep{barnett_ethical_2023} and text \citep{bommasani_opportunities_2022, weidinger_ethical_2021, liu_trustworthy_2023}, as well as combined modalities, such as text-to-image \citep{bird_typology_2023}. \citet{solaiman_evaluating_2023} provide an overview of harms from generative AI systems writ large and describe approaches to social impact analyses for each identified harm area. In our overview, we include both established and emerging risks. Established risks are defined by observed instances of harm, such as representation risks (e.g. \cite{luccioni_stable_2023, bianchi_easily_2023, birhane_multimodal_2021}). Emerging risks are anticipated based on the foreseeable capabilities of generative AI systems, such as increasingly persuasive content produced by generative AI \citep{matz_potential_2023, shevlane_model_2023}. 

We build on this prior literature to aggregate a single, holistic taxonomy of harm from generative AI systems. This taxonomy has six high-level harm areas: 1. Representation \& Toxicity Harms, 2. Misinformation Harms, 3. Information \& Safety Harms, 4. Malicious Use, 5. Human Autonomy \& Integrity Harms, 6.  Socioeconomic \& Environmental Harms (see \hyperref[tab:table1]{table 1}). 

\begin{table*}[!t]
  \centering
\caption{High-level overview of risks of harm from generative AI systems}
\label{tab:table1}
\begin{footnotesize}
\begin{tabular}{p{3cm}p{6cm}p{6cm}}
    \toprule
Harm area & Definition & Example
    \\
    \midrule
    \hline
Representation \& Toxicity Harms & AI systems under-, over-, or misrepresenting certain groups or generating toxic, offensive, abusive, or hateful content   & Generating images of Christian churches only when prompted to depict “a house of worship” \citep{qadri_ais_2023}  \\ \hline
Misinformation Harms                 & AI systems generating and facilitating the spread of inaccurate or misleading information that causes people to develop false beliefs & An AI-generated image that was widely circulated on Twitter led several news outlets to falsely report that an explosion had taken place at the US Pentagon, causing a brief drop in the US stock market \citep{alba_how_2023} \\ \hline
Information \& Safety Harms  & AI systems leaking, reproducing, generating or inferring sensitive, private, or hazardous information                                              & An AI system leaks private images from the training data \citep{carlini_extracting_2023}  \\ \hline
Malicious Use                  & AI systems reducing the costs and facilitating activities of actors trying to cause harm (e.g. fraud, weapons)                                 & AI systems can generate deepfake images cheaply, at scale \citep{amoroso_parents_2023}      \\ \hline
 Human Autonomy \& Integrity Harms  &  AI systems compromising human agency, or circumventing meaningful human control                                 &   An AI system becomes a trusted partner to a person and leverages this rapport to nudge them into unsafe behaviours \citep{xiang_he_2023}       \\ \hline
 Socioeconomic \& Environmental Harms & AI systems amplifying existing inequalities or creating negative impacts on employment, innovation, and the environment                & Exploitative practices to perform data annotation at scale where annotators are not fairly compensated \citep{stoev_experiencing_2023}  \\ \bottomrule
\end{tabular}
\end{footnotesize}

\end{table*}

We present a high-level overview of this taxonomy below (\hyperref[tab:table1]{table 1}), with examples of how these risks may manifest in modalities other than text. A more detailed breakdown of each risk area is provided in \hyperref[tab:table2]{appendix section A.1}.

\subsubsection{Multimodality raises new evaluation challenges}

While none of the higher-level risk areas are new in multimodal as opposed to text-based generative AI systems, the specific ways in which they may manifest are likely to differ between modalities. For example, violent or sexually explicit content has a greater “shock factor” in image modalities than in text. Multimodal models may also introduce novel evaluation challenges. For example, consider a text-to-image AI system that produces images based on text input. \citet{hutchinson_underspecification_2022} argue that text often underspecifies context such that an image of a “wedding”, for example, will necessarily include certain objects and cultural contexts, regardless of whether these concepts are articulated in the accompanying text. It may be easier for an AI system to hedge or give pluralist output in text than in images. 

Risks may also be compositional, i.e. manifest through the very combination of output across modalities. For example, pairing the caption “these smell bad” next to an image of a skunk is not harmful, but the same caption next to an image of a group of people may constitute harassment \citep{kiela_hateful_2021}. Similarly, an innocuous video of military training exercises combined with audio describing the invasion of a country risks creating an instance of misinformation \citep{vincent_republicans_2023}. Generative AI systems may also perpetuate stereotypes in ways that are highly dependent on domains – for example, by overrepresenting nude females as compared to nude males in the context of synthetic music videos. Detecting these harms may build on existing methods but is likely to require novel context-sensitive evaluation approaches. Though risk evaluation will likely draw from lessons learned evaluating models in single modalities, novel evaluations that enable a holistic view across modalities are required to capture risks in multimodal AI systems.

\subsection{Mapping the landscape}

We now present the results of a large-scale review of existing benchmarks to assess risks of harm from generative AI systems. To write this section, our group of co-authors and reviewers assembled an overview of all sociotechnical safety benchmarks and evaluation methods known to this group up to 10 October 2023. 

Included are academic papers or online reports that meet two criteria: they constitute an evaluation, and they have been applied to a generative AI system. An evaluation is defined as either a set of model inputs, such as a dataset, and a metric; or the application of a method (e.g. red teaming a specific AI system or a human–computer interaction study). It has been applied to a generative AI system if the publication describes results from its application to a generative AI system. Note that evaluations that may be applicable to generative AI systems but have not been applied to such systems yet were not in scope for this review.

All submitted evaluations were coded based on output modality – whether they evaluate text, image, audio, video, or multiple modalities. They were then coded based on what risk area they cover, based on the above taxonomy. Finally, they were coded by layers of evaluation in our three-layered framework (1-capability, 2-human interaction, 3-systemic impact).

We release an evaluation repository of all included evaluations as an open resource \href{https://dpmd.ai/46CPd58}{here}. The contents of this survey evaluation repository are presented in an overview figure (\hyperref[fig2]{figure 3.1}). 

\subsubsection{Limitations}

While great efforts were made to conduct a large-scale review of existing evaluation approaches, we do not assume that this mapping is comprehensive. Our approach is further limited by not considering input modality: our coding is based on output modality. Future mappings may distinguish between input modalities for a more fine-grained analysis (e.g. mapping ‘image-to-text’ evaluations distinctly from ‘text-to-text’ evaluations). Finally, our mapping is a snapshot of a moment in time. In the future, it may be conducive to a thriving ecosystem on sociotechnical evaluations to expand the evaluation repository into a living resource that evaluation developers can add their methods to.

\subsection{Evaluation gaps}

Inspecting the state of safety evaluations applied to generative AI systems reveals three high-level gaps: 

\begin{enumerate}
    \item \textbf{Coverage gap: Evaluations for several risks are lacking.} Coverage of ethical and social risk evaluation overall is low. Several gaps exist where there are few or no evaluations to assess a given risk area.
    \item \textbf{Context gap: Human interaction and systemic evaluations are rare.} Existing evaluations cluster in the text modality, with fewer evaluations available for audio, image, video, or combinations of modalities. This presents a challenge for evaluating social and ethical risks in other modalities. 
    \item \textbf{Multimodal gap: Evaluations are missing for multimodal AI systems.} Most evaluations of social and ethical harms that were identified cluster at the layer of capability evaluations. We now discuss these observations in turn.
\end{enumerate}

First, we observe that evaluations are scarce for several previously identified risks from generative AI systems. This lack of coverage is particularly pronounced for information and safety harms, human autonomy and integrity harms, and socioeconomic and environmental harms. While the number of available evaluations is insufficient to assess coverage of a risk area, the absence of evaluations is a clear signal that the given risk area cannot currently be evaluated in generative AI systems.

More detailed inspection of the evaluation repository indicates that the lack of coverage extends beyond these three risk areas: even where evaluations exist, they do not cover the risk area comprehensively. For example, we identified 83 evaluations of representation harms. However, these cover only a small space of representation harms – 17\% of them cover binary gender and occupation bias,\footnote{Six of the fourteen focused exclusively on gender and occupation. The rest include additional demographics and stereotypes.} and 60 cover text modalities only. They also cover only a small space of the potential harm: multiple “discriminatory bias” benchmarks cover binary gender or skin colour as potential traits for discrimination \citep{cho_dall-eval:_2023, mandal_multimodal_2023}. These evaluations do not cover potential manifestations of representation harms along other axes such as ability status, age, religion, nationality, or social class. In sum, further evaluations are needed to cover ethical and social harms, including plugging more nuanced gaps in risk areas for which some evaluations exist.

\begin{figure}[h]
    \centering
    \caption{Evaluations per layer. Human interaction and systemic impact evaluations to assess generative AI system safety are rare.}
    \label{fig3}
    \includegraphics[width=6cm]{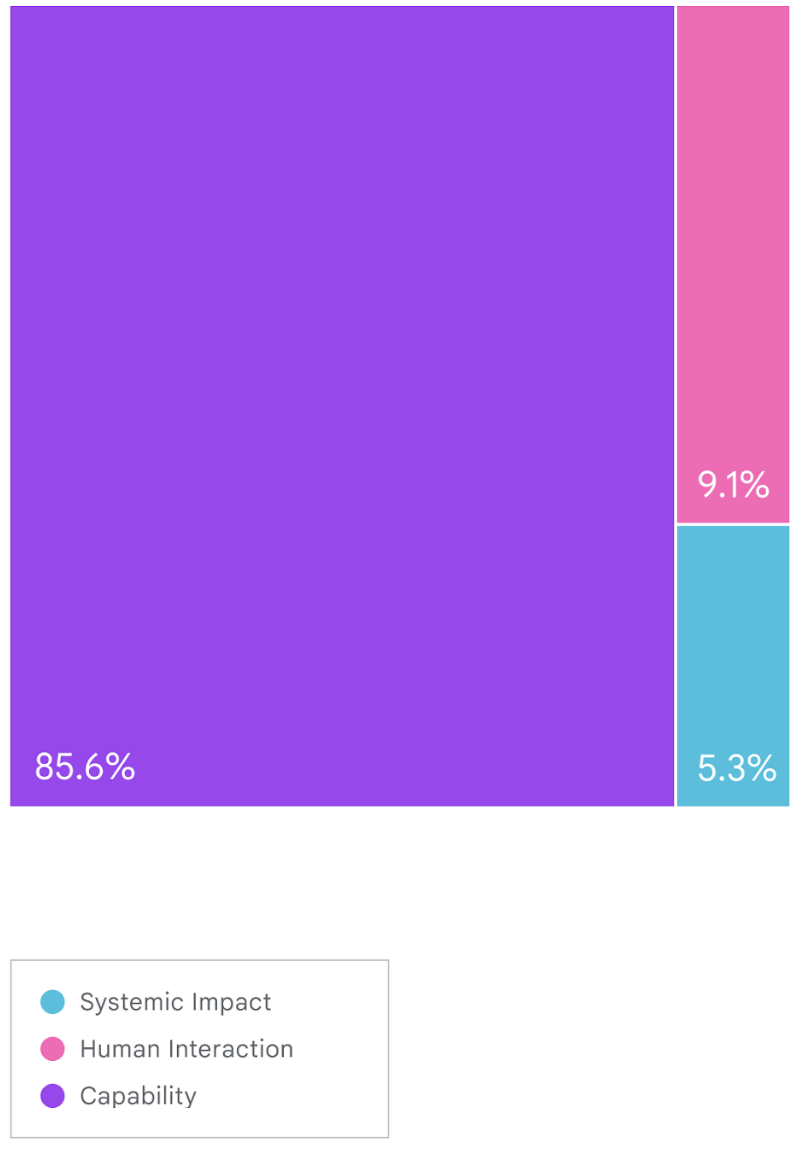}
\end{figure}

Our second main observation is that insofar as evaluation tools exist to address risks from multimodal generative AI, they are mainly clustered at the capability layer. More detailed inspection of the repository indicates that evaluations focus particularly on AI system outputs and to a lesser degree on available training data. This clustering of evaluation at the capability layer is reflective of, and likely partially driven by, the evaluations that have recently been performed and disclosed as part of large generative AI system announcements, which primarily focus on capability evaluations \citep{glaese_improving_2022, anthropic_model_nodate, openai_gpt-4_2023, mishkin_dalle_2022, touvron_llama_2023, anil_palm_2023}. 

While a capability-focused approach provides important indications as to potential downstream harms, it does not account for contextual factors that co-determine risks of harm (see \hyperref[section2]{section 2}). Capability evaluation is a core piece of safety evaluation, but it must be complemented by further analyses that add layers of relevant context. As a result, further work is needed to expand sociotechnical evaluations at the human interaction layer and at the system layer. 

\begin{figure}[h]
    \centering
    \caption{Evaluations per layer and modality. Most (75\%) of all evaluations target text output.}
    \label{fig4}
    \includegraphics[width=8cm]{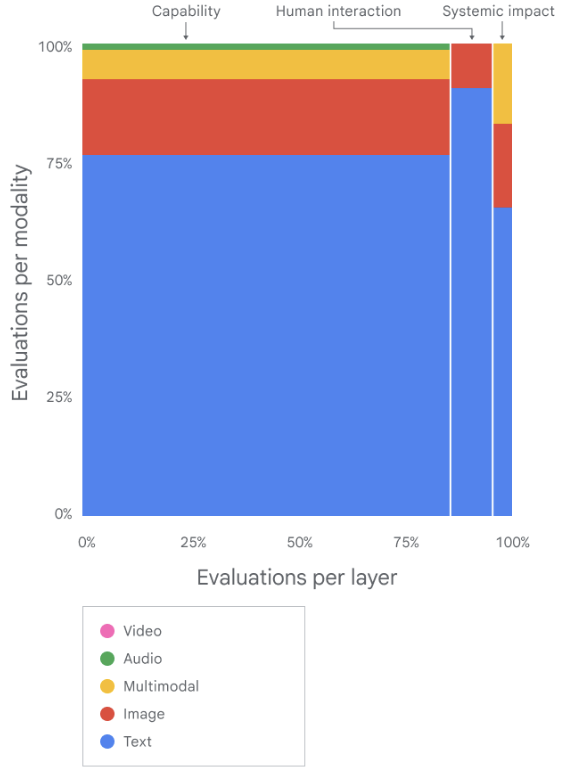}
\end{figure}

Our third observation is that the vast majority of evaluations exclusively assess text. Few evaluations exist for image outputs or combinations of text and image, and evaluations of audio or video modalities are scarce. There are only four publicly documented evaluations targeting audio and we did not find any evaluations targeting video.\footnote {Note that there are evaluations for harms arising in video that have not been applied to generative AI systems and so did not satisfy the inclusion criteria here (e.g. \citet{das_hatemm:_2023, wu_detection_2020, ashraf_youtube_2022}).} This may in part be a result of historical contingencies: generative AI systems that output text saw rapid, widespread adoption, which may have triggered proportionately more research into ethical and social risks and corresponding evaluations. 

Generative AI systems that produce compelling audio including voice and music already exist \citep{oord_wavenet:_2016, dhariwal_jukebox:_2020, borsos_audiolm:_2023, agostinelli_musiclm:_2023, huang_noise2music:_2023}, and video and audiovisual capabilities are steadily improving \citep{du_conditional_2023}. In particular, the combination of multiple modalities – through interleaved outputs, such as articles with supporting imagery; or modalities layered on top of each other, such as audiovisual video with subtitles – creates different manifestations of harm across the six identified harm areas. As a result, assessing ethical and social harm in multimodal models requires novel evaluation approaches. (We discuss some steps toward this in \hyperref[section4]{section 4}.) 

Critically, this distribution of evaluations centring text modalities is not driven by a principled assessment of the modalities in which harm is likely to occur. Several risks have been anticipated in the audio, image, and video modalities or combinations (see \hyperref[tab:table2]{appendix section A.1}). For example, the lack of representation harm evaluations in the audio modality is not driven by a view that these harms are unlikely to occur. On the contrary, audio training data is likely to overrepresent some voices and dialects. Analogous to representation harms in text-based systems, this bias may lead generative AI systems to produce higher-quality output in some voices and dialects than others. Such unfair disparities across dialects is well documented in speech recognition and in speech-to-text models \citep{ngueajio_hey_2022}, but no evaluation tools exist to assess this in generative AI systems. In some cases, evaluations designed for text output can be repurposed for other modalities (see \hyperref[section4]{section 4}). However, this is limited, especially where the same risk may manifest differently across modalities.

Combinations of modalities can create novel risks as well as compound effects. For example, misinformation has been found to be more compelling in audiovisual modalities as opposed to text \citep{hameleers_picture_2020}. AI systems that span multiple modalities may also be more vulnerable to malicious attacks aimed at getting a model to create harmful output, as fewer safety mechanisms and less exploration of vulnerabilities exist for them \citep{carlini_are_2023}. Thus, evaluations must be expanded to modalities other than text. In addition to evaluating individual modalities in isolation, they must also be expanded to assess compositions of modalities, i.e. multimodal outputs.

\section{Closing evaluation gaps} \label{section4}

Our assessment of the current state of safety evaluations of generative AI systems identified significant gaps. In this section we propose practical steps to close these gaps. These steps are tractable. Closing these gaps will require work. In addition, it may involve clarifying roles and responsibilities, which we return to in \hyperref[discussion]{section 5}. This section is primarily aimed at practitioners and those funding or performing the construction of new evaluations.

To close identified gaps, new evaluations are needed. In part, this likely means constructing novel evaluations. The first part of this section presents the general pipeline and building blocks for constructing such evaluations. In particular, we describe how rich, multifaceted concepts of harm can be made measurable through the process of “operationalisation” (see \hyperref[operationalisingrisks]{Operationalising risks}). We then outline concrete methodologies that can be used to obtain measures of a given AI system, for each layer of evaluation (see \hyperref[selectevalmethods]{Selecting evaluation methods}).

In addition to constructing novel evaluations, it may be possible to extend existing evaluations to generative AI systems. The second part of this section focuses on practical steps that can be taken to close the gaps in the evaluation of generative AI systems. We discuss these practical avenues and their advantages and limitations in the second part of this section (see \hyperref[practicalsteps]{Practical steps to closing gaps in safety evaluation}).

\subsection{Operationalising risks}
\label{operationalisingrisks}

Evaluation is a process involving several steps: it requires first selecting a target (such as a risk of harm, e.g. “bias”); then operationalising it into a concrete metric (e.g. the association of gender and occupation, \citet{luccioni_stable_2023}); then obtaining a measurement; and finally judging the outcome. Each of these steps has technical and normative elements (see \hyperref[evalneutral]{Evaluation is never value-neutral}). In this paper, we lean on previous literature to identify target constructs – namely, a taxonomy of identified risks of harm (see \hyperref[tab:table2]{appendix section A.1}). But how to proceed from a complex, multifaceted concept such as “misinformation” to a valid, tractable measurement of this risk? The process of operationally defining risks of harm such that they can be measured is the focus of this section.

Risks of harm from generative AI systems are often latent constructs that are not directly observable via a single test or metric \citep{jacobs_measurement_2021}.\footnote{This also applies to other targets of evaluation, such as cognitive capacities or potential benefits of AI systems.} In order to measure these risks, they need to be operationally defined \citep{king_designing_2021}. Operationalisation is the process of mapping tractable, observable metrics or concepts to latent constructs. Measurement on these observable metrics is then taken to provide insight on the latent target construct. Note that operationalisation is inherently an ambiguous process. What constitutes a valid measure of harm is a contestable decision, and often metrics are iterated on and improved over time \citep{chang_inventing_2004}. Operationalisation may also constitute normative trade-offs – for example, on how a single performance metric should weigh false positives against false negatives. Operationalisation of complex constructs creates various pitfalls that can result in invalid measurements (see \hyperref[operationalisingrisks]{Operationalising risks}).

In \hyperref[section2]{section 2} above, we argued that risks of harm from generative AI systems cannot be comprehensively assessed at a single layer of evaluation. Rather, complementary evaluation at all three layers is needed for a full evaluation. Thus, we propose operationally defining harm constructs at each of the three layers of evaluation. 

Specifically, this requires mapping metrics or concepts that are observable at a given layer to the latent harm construct. Different aspects of a risk can be measured at each layer. Correspondingly, different metrics can be mapped to a given risk per layer. Metrics can range from single, observable, narrow metrics (e.g. the FID score to assess the quality of a generated image) to more open-ended empirical or qualitative metrics (e.g. user preferences or broader societal impact). 

For example, to operationalise the risk of information and safety harms, we may define the following metrics at each layer. At the capability layer, we assess properties of model output that indicate potential information hazards, such as the capability to output harmful biological information \citep{noauthor_core_2023, openai_gpt-4_2023}. At the human interaction layer, this risk can be operationalised in multiple ways, one of which might be the likelihood of people unintentionally following instructions to assemble dangerous compounds in different contexts, e.g. where the generative AI system is used as a laboratory assistant. At this layer, the risk likelihood may also be measured via the friction that people encounter when intentionally trying to apply dangerous AI capabilities to malicious ends. Finally, at the systemic impact layer, this risk can be assessed via modelling potential distribution mechanisms of novel biohazards created based on such information. (These are examples for illustration purposes; for a more in-depth example, see the operationalisation of misinformation harms in our \hyperref[casestudy1]{Case study: Misinformation}.)

\subsubsection{Ensuring validity}
\label{ensuringvalidity}

Some information inevitably gets lost when operationalising complex constructs such that they can be measured – translating risks from AI systems into narrow metrics and tests is fraught with ambiguity \citep{wagner_measuring_2021}. This loss compromises the validity of a measure. There are different ways in which validity may be compromised. We briefly canvass these and outline approaches to mitigating validity concerns.

Tests often do not measure precisely what they seek out to measure: they may capture only a subset or part of the target construct (internal validity), or may capture the phenomenon fully in a given instance but not allow extrapolation to new situations (external validity) \citep{liao_are_2021}. 

AI capability evaluation in particular has been criticised for relying on overly narrow operational definitions of complex harms, leading to both internal and external validity failures \citep{raji_ai_2021, liao_are_2021}. Operational definitions must be arrived at carefully and deliberately, or they risk yielding misleading results. For example, one study found that the risk of harmful stereotyping in language modelling had been operationalised as the association of word pairs, but only some of the referenced word pairs were actually harmful and others were innocuous, such as the word pair “Norwegian” and “salmon”. As this operationalisation included instances that were not harmful, the validity of the resulting metric and what it can say about harmful stereotyping was fundamentally called into question \citep{blodgett_stereotyping_2021}. Similar validity failures have been exposed in other evaluation approaches, particularly in narrow tests such as automated benchmarks \citep{schlangen_language_2019, rauh_characteristics_2022}, which we return to later in this section (see \hyperref[selectevalmethods]{Selecting evaluation methods}).

To mitigate such validity issues, multiple approaches can be taken. Specific validity challenges for individual methods are described in detail in \hyperref[selectevalmethods]{Selecting evaluation methods} below. Here, we outline general best practices to assure the validity of a given evaluation:
\begin{itemize}
    \item \textbf{Grounding the operationalisation of a risk of harm.} An evaluation can, for example, be grounded in a literature review of a given harm, or in human annotation or examples curated by experts. To stress-test definitions and operationalisations, invite diverse perspectives and multiple lenses onto the same risk of harm. Participatory, expert-led, and interdisciplinary approaches can be helpful here (e.g. \citet{narayanan_gpt-4_nodate}).
    \item \textbf{Documenting and signposting limitations of a given evaluation.} As risks of harm are latent concepts, no single operationalisation captures them in their entirety. By making choices on how to operationalise a given risk explicit and documenting them, others can better interpret results and identify limitations \citep{noauthor_perspective:_2023, raji_ai_2021}. 
    \item \textbf{Cross-validating an operationalisation by comparing results from different evaluations of the same concept.} If the results do not align, this indicates areas in which metrics operationalise a harm in divergent ways (e.g. \citet{goldfarb-tarrant_intrinsic_nodate}).
    \item \textbf{Making results interpretable.} This may include aggregating multiple results into a single, overall result that captures multiple facets of a harm. However, collapsing multiple tests into a single result requires care as it can also make it harder to identify validity failures of individual items.
\end{itemize}
\subsection{Selecting evaluation methods}
\label{selectevalmethods}

Once a risk of harm is operationally defined, appropriate methods must be selected to obtain these measurements. Often, the selection of evaluation methods is intimately entwined with defining the metrics. In this section, we describe available methods for measuring risks of harm from generative AI systems at each of the three layers of evaluation. For each method, we provide examples of sociotechnical evaluations, and discuss methodological limitations in \hyperref[appendixa2]{appendix section A.2}.

\subsubsection{Capability evaluation methods}

To assess model capabilities, practitioners may leverage \textit{automated evaluations} that assess performance against fixed datasets or tasks. Alternatively, \textit{human annotation} can evaluate AI system capabilities against specified goals or failures, such as whether an image includes violent images. Human data annotation can be used to develop novel automated evaluations. Capability testing can be \textit{adversarial}, whereby humans or automated tools may probe a model to identify pathways that lead to failure modes. Such adversarial probing can be quite exploratory and may, in some instances, surface unexpected risks or failure modes.

Evaluation methods at this layer can be grouped as follows. We provide detailed descriptions, examples, and a discussion of limitations in \hyperref[appendixa21]{appendix section A.2.1}:
\begin{itemize}
    \item \hyperref[humanannotation]{Human annotation}
    \item \hyperref[benchmarking]{Benchmarking}
    \item \hyperref[adversarialtesting]{Adversarial testing}
\end{itemize}

\subsubsection{Human interaction evaluation methods}

This layer centres the experience of humans interacting with AI systems. Evaluation at this layer always requires human participants, as their experiences and effects or externalities on human interactants are the subject of study.\footnote{It has been proposed to simulate human participants in social science research (e.g. \cite{argyle_out_2023, dillion_can_2023}), but these methods are in their infancy – i.e. in the early, exploratory stages. They cannot therefore be relied upon for robust information to underpin responsible decision-making in AI system development.} The extent to which AI systems influence or shape human preferences and behaviours can be assessed via \textit{behavioural experiments}. These experiments can bring general mechanisms and effects into focus. Assessing the consequences of specific features, use cases, or application domains requires user \textit{research}. User studies can also assess how people actually attempt to use a generative AI system, as contrasted with the use case intended by designers. Whether an AI system functions across domains and how it performs for different user groups is core to a range of social and ethical risks, and can be assessed through user testing. While behavioural experiments and user testing require some abstraction from real-world use, passive monitoring of how people use deployed systems can provide insights on downstream effects in real-world contexts. Mixed-methods approaches that integrate different sources of data, such as behavioural observations and survey data, often provide the most robust results. Some AI system impacts on users or interaction effects may manifest only over the course of prolonged or frequent interaction; detecting these requires longitudinal designs of any of these research methods that evaluate human–AI interaction over time.

Evaluation methods at this layer can be grouped as follows. We provide detailed descriptions, examples, and a discussion of limitations in \hyperref[appendixa22]{appendix section A.2.2}:
\begin{itemize}
    \item \hyperref[behaviouralexperiments]{Behavioural experiments}
    \item \hyperref[userresearch]{User research}
    \item \hyperref[passivemonitoring]{Passive monitoring of human use}
\end{itemize}

\subsubsection{Systemic impact evaluation methods}

At the system layer, evaluation methods target the emergent effects from interactions within the sociotechnical system of which an AI system is part. This includes \textit{staged release} or \textit{pilot studies} and \textit{ex-post impact assessments} that assess the impact of AI systems on the institutions, societies, economy, and natural environments in which an AI system is embedded. Such evaluation may track broad indicators or constitute specific case studies from which broader effects are extrapolated. System evaluation also includes \textit{forecasts and simulations} to anticipate downstream harm and to identify pathways by which risks of harm may manifest. Mixed-methods approaches that combine these methods can yield more comprehensive results.  

Evaluation methods at this layer can be grouped as follows. We provide detailed descriptions, examples, and a discussion of limitations in \hyperref[appendixa23]{appendix section A.2.3}:
\begin{itemize}
    \item \hyperref[stagedrelease]{Staged release and pilot studies}
    \item \hyperref[impactassessments]{Impact assessments}
    \item \hyperref[forecasts]{Forecasts and simulations}
\end{itemize}

\subsection{Practical steps to closing the multimodal evaluation gap}
\label{practicalsteps}

So far, this paper has laid out a principled approach to implementing comprehensive safety evaluations for generative AI systems. Here, we focus on the sociotechnical evaluation gap and a framework and methods to close it. We propose some tactical steps and “quick wins” that can be taken in conjunction with establishing a more comprehensive evaluation approach. We then discuss limitations to these tactical approaches. 

\subsubsection{Repurposing evaluations for new modalities}

One way to address gaps in the evaluation landscape is to repurpose components of existing evaluation methods. Through repurposing, tools and evaluations developed for other use cases may be applied to the evaluation of generative AI systems. 

Repurposing and reusing datasets and tasks is a common approach in machine learning research and has been widely documented \citep{koch_reduced_2021, bommasani_ecosystem_2023}. For example, Winogender \citep{rudinger_gender_2018} and Winobias \citep{zhao_gender_2018} were developed as benchmarks to address the specific problem in language modelling of coreference resolution. These benchmarks are now commonly used to assess “bias” in large generative AI systems, as they quantify the association of gender and occupation in text output. They were also used as inspiration for probing generative AI systems that produce images (DALLE2 system card). Interestingly, it seems that the narrow operationalisation of a broad harm in one modality – such as operationalising bias as associations of gender and occupation in evaluations of text \citep{rudinger_gender_2018, zhao_gender_2018} – has influenced the operationalisation of the same broad harm in other modalities, as prominent image-based evaluations of generative AI systems also assess bias through associations of gender and occupation \citep{luccioni_stable_2023, naik_social_2023}. 

The way in which evaluations or their components propagate can be subtle: for example, a sentiment bias evaluation that was introduced in 2019 by \citet{huang_reducing_2020} was cited in the GPT3 paper for a modified sentiment bias analysis. In their 2021 paper presenting Gopher, \citet{rae_scaling_2022} conducted the same analysis but used an expanded set of prompts. Most recently, PaLM2 drew on the Gopher prompt set for a multilingual toxicity analysis. This practice of reuse is especially acute where AI system developers are working on tight timelines in fast-moving research domains, as is the case with generative AI.

However, this approach must be pursued with great caution. While repurposing saves work and can create common standards, applying an evaluation or classifier out of its intended context presents important trade-offs, such that repurposing, if done poorly, may create more harm than good \citep{selbst_fairness_2019}. Another example is hate speech classifiers, which are typically trained on dialogue data between two people – for example, on social media. There are very few datasets on human–AI interaction and the ways in which hate speech may emerge in that context. To determine if and when an evaluation should be reused, practitioners may consider its provenance, identify how the original context and purpose aligns with the new usage, and understand what norms are being perpetuated by its reuse. Because risks of harm are contextual, understanding the difference between the original and updated context will uncover the gaps in the new use case, including validity issues (see \hyperref[operationalisingrisks]{Operationalising risks}). 

Rather than simply repurposing existing evaluations to assess risks in other modalities, these tools may be used as a starting point for refinement, or as a guiding analogy for constructing new evaluations. Existing methods for evaluating these risks may be a useful template that can be refined, or replicated in a way that matches novel capabilities and provides meaningful evaluations of generative AI systems.

\subsubsection{Transcribing non-text output for text-based evaluation}

Another way to address the uneven distribution of evaluations across modalities is to translate outputs from one modality into another, to enable evaluation using existing methods. This may be attempted through transcribing content from images, video, or audio output such that the transcript can then be evaluated using text-based evaluation tools. For example, automatic speech recognition tools can be leveraged to transcribe speech into text or an image captioning system can be used to caption a generated image (e.g. \cite{wiles_discovering_2023}). Similarly, video can be split into a series of images to enable image-based evaluation. 

This approach is a valuable and tractable first step in evaluating risks of harm in non-text modalities. However, through the process of transcription, some information inevitably gets lost and thus evades evaluation. For example, in speech, prosody (the way in which something is said, e.g. with sarcasm) carries information about meaning but might not be translated well into text \citep{wilson_relevance_2006}. Similarly, generating synthetic audio in the voice of a particular person may create appropriation or defamation harms that would not be detected by transcribing what was said and analysing the text. 

Pitfalls of the transcription approach also stem from the fact that methods to translate between modalities may be error-prone \citep{rohrbach_object_2019, ramesh_beach_2022}, sometimes in systematically biased ways \citep{ngueajio_hey_2022, wang_measuring_2022}. Such errors can propagate through the harm analysis – for example, if an image-captioning system is biased toward masked athletes as “male”, evaluation of image captions may indicate a different gender bias than is present in the images that are the target of evaluation. In sum, while transcription approaches are a promising first step, these methods are limited, require quality checks, and must be complemented by evaluation methods that are calibrated to the output modality directly.

\subsubsection{Model-driven evaluation may fill gaps}
\label{modeldriven}

Pre-trained generative models themselves are being used as evaluation tools because of the flexibility and generality they offer. Language models have been used to procedurally generate adversarial prompts to elicit harmful outputs from other language models \citep{perez_red_2022} and to critique model outputs as part of mitigation \citep{bai_constitutional_2022, wiles_discovering_2023}. GPT-4 was fine-tuned using a copy of the same model, prompted with a safety rubric \citep{openai_gpt-4_2023}. Advantages of these approaches are that they can use existing AI systems with little or no adaptation to the task, using a prompt or fine-tuning to guide the AI system to perform the desired benchmark or red teaming task. These methods are easier to use than developing a static benchmark from scratch. As such, they offer a way to respond more rapidly to novel risks and to cover the combinatorial space of risks and modalities. They can also mitigate the drawbacks of evaluations using human raters, which are typically costly and slow, and put the raters themselves at risk.

However, AI systems as evaluation tools face additional limitations. They rely on proprietary AI systems that may not be accessible to those performing an evaluation. These AI models are also updated over time and generate prompts stochastically, which may adversely impact the reproducibility of this approach. In addition, generative models may have biases and behave in unexpected ways, which can introduce confounds or noise into the evaluation. There is a further risk of spiralling effects if AI systems from the same model “family” are used to evaluate each other, as existing biases or blindspots present in these systems can be amplified through this process. This method is also limited in the types of risks it can address: it is primarily useful for covering risks from “unsafe” outputs, rather than risks from what the AI system omits or is not capable of (such as uneven or low performance). Finally, while promising, this direction of evaluation is novel and its robustness needs to be assessed. Grounding the results of these evaluations by comparing with human or other established evaluations is a critical cross-validation step to ensure this method does not fall foul of validation problems (see \hyperref[operationalisingrisks]{Operationalising risks}).

\section{Discussion} \label{discussion}

\subsection{Benefits of a sociotechnical approach}

Evaluating technical components (such as the data an AI system was trained on) or AI system behaviour (such as outputs in response to prompts) is important, but insufficient, for determining whether an AI system is safe. This is for two reasons: First, potential harms from AI are felt and observed outside of technical AI system evaluations themselves. While evaluation of AI system capabilities can serve to predict risk of harm, it is a proxy for the actual downstream harm that may be experienced. Second, risks of harm can emerge from interactions between multiple factors, including technical components, human factors, and structural factors such as the broader systems in which an AI system is deployed. As these risks of harm are emergent through the interaction of these factors, context determines whether or not an AI system is safe \citep{leveson_engineering_2012}. Thus, to assess whether an AI system is safe requires evaluating these different layers of context.

In this paper, we lay out a sociotechnical, three-layered framework to evaluate the safety of generative AI systems. The benefit of taking a multilayered, sociotechnical approach is that it takes into account the context that ultimately determines, the safety of an AI system \cite{leveson_engineering_2012}.  By carefully laying out the steps toward implementing this evaluation framework, we demonstrate that a sociotechnical approach to better AI safety evaluation is insightful, needed, and tractable. 

\subsection{Roles and responsibilities} \label{rolesandresponsibilities}

\begin{figure}[h]
    \centering
    \caption{Responsibilities for conducting evaluations are shared between different AI actors. Primary responsibility depends on which actors are best placed to conduct evaluations at this layer.}
    \label{fig5}
    \includegraphics[width=8cm]{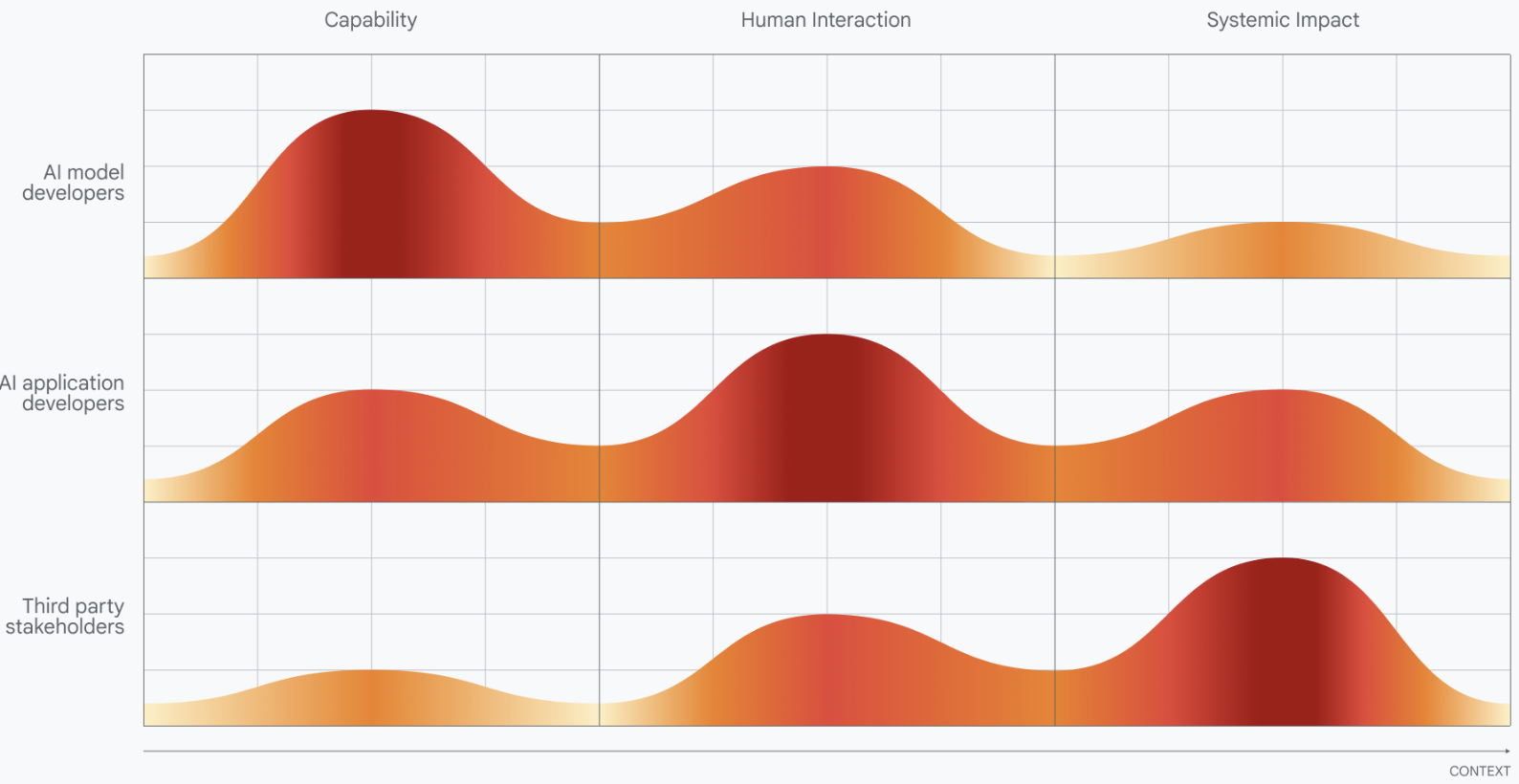}
\end{figure}

Fostering a thriving sociotechnical evaluation ecosystem requires clear roles and responsibilities amongst the various AI actors.\footnote{The OECD defines AI actors as “those who play an active role in the AI system lifecycle, including organisations and individuals that deploy or operate AI” \citep{organisation_for_economic_co-operation_and_development_artificial_nodate}.} This includes AI developers, vendors, and product developers, as well as public sector and civil society stakeholders. While the responsibility for conducting comprehensive evaluations to determine the safety of AI systems is shared between private and public stakeholders, different actors will be better placed to perform an evaluation for a given layer due to factors such as in-house expertise, proprietary infrastructure, or established practices (voluntary or statutory). More often, all actors have some responsibility to ensure comprehensive evaluation of risks of harm across each of the layers (\hyperref[fig5]{figure 5.1}).

Given their degree of knowledge and autonomy over what they are building, AI developers have the primary responsibility for conducting sociotechnical evaluations pertaining to the AI system \citep{owen_organisational_2021, stilgoe_developing_2013, dignum_responsible_2019}. There are good reasons for others, such as independent third-party auditors, also to perform capability evaluations \citep{raji_outsider_2022}. However, AI developers have a responsibility to ensure that the capabilities of the systems they build have been evaluated for safety. AI application developers are best placed to evaluate human interaction effects, including functionality across application domains and groups, and possible externalities. As application developers modify some technical components, such as filters, or may be able to adapt an AI system to a specific use case (“fine-tune”), they are well placed to conduct some evaluation on technical capabilities of these parts. Further, application developers may have proprietary access to data on how an AI system is used by consumers, placing them in a key position to conduct systemic impact evaluations using this data. Note that the roles of AI model and AI application developers often converge in practice, where organisations who develop an AI system also deploy it or make it available for user-facing products. In these cases, it may be the same organisation that bears responsibility for capability and human interaction evaluation. AI model developers and AI application developers may be private, academic, or public actors.

Third-party stakeholders – such as governments, civic interest groups, groups representing technology users, or private organisations – are often best positioned to perform evaluations of systemic impact. These actors can leverage specialist knowledge in a given domain (e.g. financial, environmental, public health) where risks may arise, both for foresight and evaluation. Public actors such as governments further have the responsibility of ensuring public safety, which is anchored at the systemic impact layer. Systemic impact evaluation also often lasts over long periods of time, which maps onto the responsibilities of public actors to keep a long-term view of public safety. Through systemic impact evaluations, third-party stakeholders such as governments or regulators may also have access to public data that can provide the basis for systemic impact evaluations. However, third-party stakeholders may also be well placed to evaluate risks from AI applications in specific domains, particularly in high-stakes contexts, at the human interaction layer (e.g. \citet{noauthor_ai_2021}). Capability and human interaction testing of proprietary, unreleased systems may in some cases require novel infrastructure, incentive structures, and standardised evaluation approaches for evaluators and developers to coalesce around, as well as reliable safety assurances. 

Note that the three layers are not contingent on each other: rather, evaluation at all three layers can be conducted in parallel. To some extent, the layers track AI system development from basic capabilities, to applications and user testing, to broader deployment. However, this does not mean that evaluation at these layers follows a chronological sequence. Rather, there are evaluations at each layer that can be performed at any point of AI system development. To list just a few examples, such prospective evaluations include assessing training data (capability layer), psychological mechanisms at play (human interaction layer), and economic impact of comparable technologies (system layer). (For more examples, see \hyperref[selectevalmethods]{Selecting evaluation methods}.) Evaluations at each layer can be performed simultaneously and asynchronously. 

\subsection{Limits of evaluation}

At the same time as expanding sociotechnical evaluations, it is important to keep a clear mind on the limits of what evaluation can provide, such that evaluations can be embedded in a broader sociotechnical approach to ensuring safe AI systems. Evaluation, as noted above, is a core component of responsible innovation: it links up foresight and observed accidents with actionable responses such as mitigation and responsible decision-making. Nevertheless, evaluations are not a panacea for ensuring safe AI systems. In this section we outline these limits and challenges of evaluation.

\subsubsection{Evaluation is incomplete}

Evaluation cannot catch all potential risks of harm, for several reasons. First, evaluation necessarily and inherently covers only a subset of all possible manifestations of risks of harm \citep{bergman_representation_2023}. What is included depends on pragmatic and normative considerations, such as what is tractable, anticipated, and prioritised (see \hyperref[evalneutral]{Evaluation is never value-neutral}). Areas for which simply no evaluation exists, or where it is not technically or otherwise viable to implement these evaluations \citep{perlitz_efficient_2023}, remain unevaluated. This means that some safety-relevant aspects – for example, failure modes specific to particular user groups, application domains, or intersections of such factors – are outside the purview of evaluation. In addition, unknown and other unanticipated failure modes are, by definition, not tested for and may go undetected. 

The incompleteness of evaluation is particularly apparent in the context of “general-purpose” generative AI systems, whose downstream application or user base is not yet defined or understood. An often-cited ambition in the innovation of generative AI systems is to develop “general-purpose technologies” that could be applied to a wide range of potential tasks and environments (e.g. \citet{bubeck_sparks_2023}, though see also \citet{raji_ai_2021}). Indeed, generative AI systems have been likened to general-purpose technologies such as steam engines and office automation \citep{acemoglu_power_2023}. This supposed open-endedness of AI systems can make it difficult to identify the contexts – such as applications, user groups, or institutions – in which AI system safety should be evaluated. 

One way to address this tension in practice is to define hypothetical applications of “general-purpose technologies” and to evaluate them in these contexts. This can, for example, take the form of identifying “critical user journeys”, i.e. mapping a series of steps users may take using a product to achieve a desired outcome \citep{arguelles_critical_2020}. Following a precautionary approach, such hypothetical use case mapping may first focus on high-risk applications. Such early evaluation based on hypothetical use cases cannot replace downstream evaluation of actual use cases; rather, it serves to highlight potential risks and must be complemented with monitoring of real-world impacts. The risk profiles and thresholds of what constitutes “acceptable” model performance may differ between different downstream applications or user groups, requiring more rigorous evaluation in some cases than in others. 

In some cases, evaluation may further be incomplete because it would be inappropriate or problematic, or create a disproportionate burden to perform evaluations. For example, measuring sensitive traits to assess usability across demographic groups may place communities at risk or sit in tension with privacy, respect, or dignity (e.g. \citet{wenger_finding_2022, wolff_fairness_2010}). Characteristics or qualities that are essentially contested or fundamentally fluid (e.g. ethnicity, sexual orientation, or gender identity) may be reified through evaluations that bin these into categories \citep{keyes_counting_nodate, lu_subverting_2022, tomasev_fairness_2021}. Finally, while it is important to include different communities in qualitative and other evaluation approaches, evaluation may not be desirable to the community represented \citep{denton_genealogy_2021, noauthor_please_nodate}, either due to the burden (e.g. time and labour costs) of participation or, for example, if inclusion within the scope of the evaluation means being surveilled \citep{brunton_obfuscation:_2016, keyes_counting_nodate, noauthor_please_nodate, bedoya_big_2014, hassein_against_2017}. 

Further reasons for the incompleteness of evaluation relate to the fact that some risks of harm are exceedingly difficult to operationalise and measure accurately. For example, whether an AI system promotes discriminatory race-based stereotypes is a focus of safety evaluation. However, treating social constructs such as race as \textit{fixed attributes} in evaluation may create a distorted view of the actual differential impacts on different racial groups and intersectionalities, such as with social class \citep{hanna_towards_2020}. Similarly, some harms are particularly difficult to trace, even in hindsight – such as the long-suspected and now-evidenced link between social media and teenage eating disorders \citep{noauthor_social_nodate}. Where effect sizes are small and causal mechanisms poorly understood, evaluation may fail to detect risks that it seeks to measure. This may also affect the detection of potential emergent capabilities that may only become observable as an AI system reaches a certain scale. In addition, overall impacts of an AI system on complex notions such as welfare are difficult to measure because the target construct itself (welfare) is difficult to establish. Especially where effects are distributed and interact with other factors such as user vulnerabilities, it can be difficult to establish hard findings. Long-term and mixed-methods approaches, including initial qualitative work, can help reduce these limitations and shed light on potential subtle or highly indirect effects. 

In sum, even with best efforts, there will always be harms of particular kinds or in particular contexts that are not evaluated. This is why evaluation must be complemented with effective governance mechanisms for evaluating remaining uncertainties prior to AI system release, with post-deployment monitoring including logging observed incidents \citep{noauthor_welcome_nodate} and with well-functioning and swift recourse mechanisms for people who experience or detect harm. It is important that AI systems are flexibly designed such that new insights can be translated into fixes, such as via system updates. Given the pre-deployment evaluation gaps, organisations deploying AI systems require adequate governance infrastructures that can respond to detected risks with mitigations, by delaying or stopping the deployment of an AI system or by suspending an already-deployed system until concerns are resolved.

Generative AI systems pose significant risks for individuals, communities, and society, and failing to detect and mitigate such risks can have serious consequences \citep{xiang_he_2023}. This is why it is critical to ensure that evaluation is prioritised and that complementary mechanisms exist to uphold AI system safety to cover the gaps that are inherent limitations to evaluation.

\subsubsection{Evaluation is never value-neutral}
\label{evalneutral}

Evaluations are inherently value expressions of those who conduct them: they always require a decision on what is valued \citep{bowker_sorting_2000}. It is widely understood and expressed in the sociotechnical literature that AI systems are not merely mathematical constructs but sociotechnical and political entities, with inherent value systems embedded in the choices made by designers with respect to how to create and implement the model \citep{barocas_fairness_2019, birhane_values_2022, gururangan_whose_2022, raji_ai_2021, sambasivan_re-imagining_2021, scheuerman_datasets_2021, suresh_framework_2021}. The same extends to evaluations of an AI model or system. 

There are normative decisions throughout the evaluation construction and implementation process that cannot be avoided. No evaluation can cover every circumstance and dimension of what can be evaluated \citep{bergman_representation_2023, raji_ai_2021}. Thus, designing an evaluation involves choices – made either deliberately or implicitly – on what to prioritise and what to discard. First, selecting a target to evaluate requires a normative judgement on what harms are important or relevant to measure \citep{kalluri_dont_2020, mohamed_decolonial_2020}. Operationalising the harm further requires normative decisions on what task is most valuable for the system to perform highly on, what high performance looks like, and where or to whom it is most valuable/optimal for the benefits of the system to accrue. After this process, what remains within scope of the evaluation is what is prioritised, and these decisions inherently express value. 

Furthermore, operationalising a harm construct into a metric necessarily bakes in certain assumptions. For example, making a commitment to a test and a metric – e.g. that “social biases'' can be measured via associations of gender and occupations – is a normative judgement on where harms are likely to occur and which particular aspects of a harm are relevant and tractable \citep{luccioni_stable_2023}. These normative decisions are all the more significant, as they tend to have a sticking effect that propagates (see \hyperref[practicalsteps]{Practical steps to closing the multimodal evaluation gap}).

Assessing whether a model meets expectations prior to or post deployment requires a normative evaluation of whether some measurement expresses performance that is “good”, “bad”, “safe enough”, etc. (see \citet{bakalar_fairness_2021}). For such thresholds to be legitimate, they need to arise from adequate institutions or processes, such as expert groups, democratic institutions, or fair and inclusive deliberation processes that centre groups that may be affected by these AI systems. The thresholds of what constitutes “acceptable” model performance may differ across use cases, contexts, or applications.\footnote{For example, generative models may be used in an application that provides access to news content or in a creative collaborator tool. Factually incorrect outputs may be of greater concern in the former than in the latter, depending on contexts such as the user expectations and de facto uses of the model (i.e. whether people assume and rely on the model as a source of truth). While the same evaluation for factually incorrect outputs can be run on these different use cases, different thresholds of acceptable levels of model performance may be applied. This requires performing general evaluations as early as possible and transparently disclosing results such that downstream users or product developers can make informed decisions on whether the model is fit for their intended purposes.} 

Another aspect in which evaluations express value is who and whose perspectives are represented in the evaluation – for example, the English-speaking or Western world \citep{gururangan_whose_2022, devries_does_nodate, shankar_no_2017}; tech-savvy and educated ML designers; socioeconomically privileged users providing feedback on a system; or annotators that skew young, female, educated, and white \citep{ding_impact_2022}. Some harms will be missed if communities that may be affected are overlooked or dimensions of harm are ignored. Calls for greater representation of community groups is widespread and often offered as a mitigation for a broad range of fairness and sociotechnical harms (e.g. \citet{costanza-chock_design_2020, devries_does_nodate, noauthor_ai_2021, noauthor_sortition_nodate,lashbrook_ai-driven_2018,jindal_responsible_2021, pasquale_opinion_2021, suresh_framework_2021, noauthor_proposal_2021}). When a community is missing in evaluation, there is no pre-deployment evaluation of the impacts of the system on that community, which can lead to skewed decision-making that ignores potential effects on these groups \citep{bergman_representation_2023, buolamwini_gender_2018}. While greater inclusion of groups and perspectives in the evaluation can lead to better visibility of the model performance \citep{buolamwini_gender_2018}, there are arguments for adopting this approach with care \citep{bergman_representation_2023}, as, for example, an oversimplified interpretation of this notion can lead to objectification and exploitation (e.g. \citet{fussell_how_2019}).

Normative decisions on AI systems that affect large groups of people ought to be made in legitimate and accountable ways. These decisions should be dynamic such that they can be updated over time and adapted to specific application domains or locales within some range of acceptability. As a first step, identifying normative choices and documenting them can provide the basis for accountability. Second, the acknowledgement that normative choices are not merely pragmatic or technical provides a foundation for shared responsibility and public engagement on certain decisions pertaining to AI safety (e.g. what should be evaluated). Ideally, significant normative decisions should be made via deliberate processes that employ inclusive, participatory techniques \citep{birhane_power_2022}. In addition to providing legitimacy, such processes may result in more robust evaluations that better track the most significant risks of downstream harms from AI systems. Documentation of how such decisions are made serves to provide further transparency and provides insights into potential limitations of evaluations \citep{raji_ai_2021}.

\subsection{Steps forward}

For evaluation to be impactful, four conditions must be met. First, evaluations on relevant risks of harm must exist. Second, these evaluations must be conducted regularly. Third, the conduct of these evaluations must have teeth, i.e. have meaningful consequences. Fourth, the conduct of these evaluations must become increasingly standardised and independent to ensure valid evaluations over time. 

\subsubsection{Evaluations must be developed where they do not yet exist}

A review of the current state of safety evaluations suggests a pressing gap exists in the collective safety evaluation toolkit. This gap will likely be further exacerbated as increasingly capable multimodal models are released more broadly at pace. In addition, evaluations are often conducted in an ad hoc manner and too late to anticipate preventable harm. This assessment calls for a strong focus, concerted action, and shared priority to develop safety evaluation.

\subsubsection{Evaluations must be done as a matter of course}

Evaluations must be conducted during the process of AI development, to bake in ethical and social considerations from the inception of an AI system rather than imperfectly patching them on as an afterthought. In particular, evaluations should be conducted from the moment of planning a new AI system. In the context of training large generative AI systems, early indicators of risks of harm can already be obtained from analysing technical components, such as training data, or a small, raw, pre-trained model. Adequate safety testing evaluates these components to influence responsible decision-making, such as by indicating whether a training dataset is appropriate for use or whether a model training course should be continued. Early testing allows decision-makers to verifiably consider the best information available at the time of making consequential decisions. Early evaluation is also a basic tenet of safety engineering, as it supports risk mitigation and the “cost of fixing” a hazard tends to increase exponentially with time in a system’s life cycle \citep{leveson_engineering_2012}. 

Evaluation must also be a continuous practice. Evaluation is subject to Collingridge’s Dilemma, which asserts that early-stage evaluation creates higher potential to influence the direction of a technology but later-stage evaluation provides more accurate and comprehensive information \citep{collingridge_social_1982}. In other words, early evaluation is less accurate but has the capacity to shape AI development more deeply. This dilemma may be somewhat modulated in the case of software systems that can, at some cost, be updated over time and post deployment. Yet to ensure that evaluation can both shape early decision-making and provide an accurate picture of risks of harm, it must be conducted at multiple time points throughout the AI system life cycle, including by monitoring effects post deployment. 

The frequency of evaluations is subject to a further tension described by Goodhart’s Law, which asserts that a measure that becomes a target ceases to be a good measure. Evaluations that are run frequently, e.g. as a method of performance tracking for AI \textit{development}, become de facto targets over time. AI designers aim to improve AI system performance on these particular metrics. To provide valid \textit{assurances} on AI system safety, evaluations cannot be used as de facto targets. To achieve this, it is important that assurance evaluations are not shared with or reverse-engineerable by AI system developers. This, in turn, requires meaningful separation between development and assurance evaluations, and between the actors who develop and those who evaluate AI systems. These separations are overall best practice in evaluation but especially important on assurance evaluations. 

Finally, AI evaluations in practice face a “realism trade-off” between the pragmatic costs of conducting an evaluation on the one hand and the accuracy and validity of results on the other \citep{liao_rethinking_2023}. This means that evaluations sit on a spectrum between yielding high accuracy – such as longitudinal, highly localised ethnographic studies – and being automated and highly generalisable – such as automated tests and benchmarks. In other words, high-frequency evaluations are attractive to AI developers but face significant validity constraints and cannot capture the scope of complex, real-world harms \citep{raji_ai_2021, rauh_characteristics_2022} (see \hyperref[ensuringvalidity]{Ensuring validity}). Establishing a mixed-methods practice is the way forward to ensuring tractable evaluations for development, as well as obtaining the best information available to underpin responsible decision-making. 

\subsubsection{Evaluation must have real consequences}

For evaluation to be impactful, it must have real consequences. Evaluation derives its relevance from the processes and decisions into which it is meaningfully embedded. The importance given to safety evaluations shows first and foremost in the organisational resources dedicated to running and considering the results of such evaluations. Ensuring meaningful safety evaluation requires that such evaluations are conducted in good time to influence decisions rather than after the fact. It also requires that evaluation results are shared with decision-makers, whether these are internal or external to an AI organisation. Importantly, safety evaluation requires that organisational structures and incentives are put in place to perform these evaluations. This includes allocating clear responsibilities to accountable, skilled, and well-staffed teams who can build and execute evaluations, are distinct from AI developers, and can serve as accountable ethics “owners” \citep{metcalf_owning_2019}. These teams further require incentive structures to perform accurate evaluations as well as supporting infrastructure such as appropriate computational resources. 

\subsubsection{Evaluations must be done systematically, in standardised ways}

As in other fields of safety engineering, increasing standardisation and independence of safety evaluation is likely to lead to a more accountable, reliable, and safe AI ecosystem. While safety evaluation of AI systems is yet to be standardised and roles and responsibilities are yet to be assigned between different actors, it is clear that safety evaluation will play a key role in ensuring the safety of generative AI systems. Independent evaluation is not commensurate with testing conducted by AI developers: both are necessary. In particular, AI developers have the capacity to use safety evaluations as a north star to guide iterative AI system development. This is complementary to comparable, verifiable, and independent evaluations that can provide more wide-ranging assurances.

Just like audits, evaluations could be conducted by actors that are independent from AI developing organisations. This would provide added credibility as well as the structural separation that makes it easier to withhold evaluations from AI developers. Keeping evaluations secret from AI developers is key to preventing an evaluation from becoming a de facto target. In addition to withholding evaluations, independent actors should ensure that evaluations are verifiably validity-tested (see \hyperref[ensuringvalidity]{Ensuring validity}), meaningful to the real-world application of an AI system, and updated over time, both to account for changes in the AI system \citep{diaz_scaling_2023} and to avert Goodhart’s Law as described above.

\subsubsection{Toward a shared framework for AI safety}

The emergence of generative AI systems and applications has led to a renewed debate about observed risks from AI technologies that can be seen to recur in generative AI systems \citep{bommasani_opportunities_2022, weidinger_ethical_2021, luccioni_stable_2023, bianchi_easily_2023, birhane_multimodal_2021, carlini_extracting_2023}. Simultaneously, the current and future classes of generative AI systems have been claimed to possess novel capabilities that may create “extreme” risks to society, such as from disseminating dangerous information or creating novel types of cyber attacks \citep{shevlane_model_2023}. Historically, these focus areas – or ethical and safety risks – associated with AI systems have been fragmented and have constituted distinct research communities based on perceived epistemic differences and differences in timely proximity of harms \citep{prunkl_beyond_2020}. However, recent advances in generative AI systems are forcing a collapse of these epistemological silos, as these domains of risk are increasingly converging in terms of their timescales and the comparability of their underlying technology. The sociotechnical approach put forward here accommodates risks that are of concern to both research communities and it can thus serve to coordinate work between these communities on risks from generative AI systems. 

Maintaining a sufficiently calibrated mapping of potential risks from AI systems requires an acknowledgement that risks are inherently conditional on the underlying technological capability and are dynamic (evolve in nature over time). Specifically, one could consider risks evolving along a “pathway” between time and capability, where the precise manifestation of each risk area (e.g. representational harms) is altered as AI systems grow more complex and generalisable in performance \citep{chan_harms_2023}. However, across AI system capability levels, the underlying risk area remains. For example, bias and toxicity in NLP systems are known issues in the field \citep{dixon_measuring_2018} but evolved in complexity when assessed within the context of generative AI systems \citep{rae_scaling_2022, openai_gpt-4_2023-1}. Frontier or advanced AI systems will likely encourage further evaluation of generative AI systems which address both clusters of risks in concert while remaining firmly grounded in the trajectory of the system’s actual developmental path. Considering both existing risks and future capabilities will allow for mapping out potential risks more robustly. This, in turn, will serve to develop more robust mitigations and governance of these risks.  

\section{Conclusion} \label{conclusion}

In this paper, we lay out a sociotechnical approach to evaluating risks from generative AI systems. We present a three-layered framework that expands the remit from capability testing, to take into account the context in which an AI system is used and its broader impact on the structures in which it is embedded. Surveying the current state of sociotechnical evaluation, we identify significant gaps related to specific risk areas, non-text modalities, and evaluations that take into account human and broader systemic context. We provide a pragmatic roadmap on how to close these gaps. Specifically, we survey available evaluation methods and tactical approaches to extend existing evaluations. We also lay out our vision of what sociotechnical evaluation can look like in key social and ethical risk areas – misinformation, representation risks, and dangerous information. Finally, we close with a review of the limits of evaluation, normative considerations, and suggestions for a practical and tractable way forward.

\clearpage

\appendix

\onecolumn

\section{Appendix}

\subsection{Taxonomy of harm} \label{taxonomy}
\label{tab:table2}
\begin{footnotesize}
\begin{longtable}{p{3cm}p{5cm}p{5cm}}
    \toprule
    % \centering
    %   \toprule % horizontal line
  \multicolumn{1}{p{0.1\textwidth}}{Risk area} &  \multicolumn{1}{p{0.15\textwidth}}{Definition} & \multicolumn{1}{p{0.15\textwidth}}{Example} 
    % \textbf{Risk area}     & \centering \textbf{Definition} & \centering \textbf{Example} 
    \\ \hline
\multicolumn{3}{c}{\cellcolor[HTML]{14234B}\color[HTML]{FFFFFF} {Representation \& Toxicity Harms}} \\ \hline
Unfair representation                                      & Mis-, under-, or over-representing certain identities, groups, or perspectives or failing to represent them at all (e.g. via homogenisation, stereotypes) & Generating more images of female-looking individuals when prompted with the word “nurse” \citep{mishkin_dalle_2022}* \\ \hline
Unfair capability distribution                             & Performing worse for some groups than others in a way that harms the worse-off group & Generating a lower-quality output when given a prompt in a non-English language \citep{dave_chatgpt_nodate}*                                                                                                                             \\ \hline
Toxic content                                            & Generating content that violates community standards, including harming or inciting hatred or violence against individuals and groups (e.g. gore, child sexual abuse material, profanities, identity attacks) & Generating visual or auditory descriptions of gruesome acts \citep{knight_joy_nodate}±, child abuse imagery \citep{harwell_ai-generated_2023}*, and hateful images \citep{qu_unsafe_2023}                                                                                                                                                    \\ \hline
\multicolumn{3}{c}{\cellcolor[HTML]{14234B}\color[HTML]{FFFFFF} {Misinformation Harms}} \\ \hline
    Propagating misconceptions/ false beliefs & Generating or spreading false, low-quality, misleading, or inaccurate information that causes people to develop false or inaccurate perceptions and beliefs & A synthetic video of a nuclear explosion prompting mass panic \citep{alba_how_2023}*                                          \\ \hline
Erosion of trust in public information                                            & Eroding trust in public information and knowledge & Dismissal of real audiovisual evidence (e.g. of human rights violation) as “synthetic” in courts \citep{gregory_fortify_2023}±; \citep{christopher_indian_2023}*; \citep{bond_people_2023}*                                                                          \\ \hline
Pollution of information ecosystem                         & Contaminating publicly available information with false or inaccurate information & Digital commons (e.g. Wikimedia) becoming replete with synthetic or factually inaccurate content \citep{huang_generative_nodate}±                                                                                                    
\\ \hline
\multicolumn{3}{c}{\cellcolor[HTML]{14234B}\color[HTML]{FFFFFF} {Information \& Safety Harms}} \\ \hline Privacy infringement                                       & Leaking, generating, or correctly inferring private and personal information about individuals  & Leaking a person’s payment address and credit card information \citep{metz_openai_2023}*                                                                        \\ \hline
Dissemination of dangerous information                  & Leaking, generating or correctly inferring hazardous or sensitive information that could pose a security threat & Generating information on how to create a novel biohazard \citep{openai_gpt-4_2023}±  \\ \hline

\multicolumn{3}{c}{\cellcolor[HTML]{14234B}\color[HTML]{FFFFFF} {Malicious Use}} \\ \hline Influence operations                                             & Facilitating large-scale disinformation campaigns and targeted manipulation of public opinion & Creating false news websites and news channels to influence election outcomes \citep{satariano_people_2023}*; \citep{vincent_republicans_2023}*
          \\ \hline
Fraud                                                      & Facilitating fraud, cheating, forgery, and impersonation scams &  Impersonating a trusted individual’s voice to scam them (e.g. providing bank details) \citep{verma_they_2023}*; \citep{krishnan_fraudgpt:_2023}*                                                                                                                                    \\ \hline
Defamation                                                 & Facilitating slander, defamation, or false accusations & Pairing real video footage with synthetic audio to attribute false statements or actions to someone \citep{burgess_ukraine_2022}±                                                                                                                                  \\ \hline
Security threats                                           & Facilitating the conduct of cyber attacks, weapon development, and security breaches & Generating code to hack into government systems \citep{burgess_generative_2023, shevlane_model_2023}±                                                                                                                                                                                                                                        \\ \hline
\multicolumn{3}{c}{\cellcolor[HTML]{14234B}\color[HTML]{FFFFFF} {Human Autonomy \& Integrity Harms}} \\ \hline Violation of personal integrity                            & Non-consensual use of one’s personal identity or likeness for unauthorised purposes (e.g. commercial purposes) & Generating a deepfake image, video, or audio of someone without their consent \citep{hunter_ai_2023}*                                                           \\ \hline Persuasion and manipulation                                               & Exploiting user trust, or nudging or coercing them into performing certain actions against their will (c.f. \cite{burtell2023artificial, kenton2021alignment}) & A personalised AI assistant persuading someone to harm themselves \citep{xiang_he_2023}*                                                                                    \\ \hline Overreliance                                               & Causing people to become emotionally or materially dependent on the model & Skill atrophy (e.g. decreased critical thinking skills) from excessive model use \citep{bai_chatgpt:_2023}±                                                                                                              \\ \hline
Misappropriation and exploitation                              & Appropriating, using, or reproducing content or data, including from minority groups, in an insensitive way, or without consent or fair compensation & Training an image-generating model on an artist’s work without their consent \citep{chen_artists_2023}*                 \\ \hline
\multicolumn{3}{c}{\cellcolor[HTML]{14234B}\color[HTML]{FFFFFF} {Socioeconomic \& Environmental Harms}} \\ \hline Unfair distribution of benefits from model access          & Unfairly allocating or withholding benefits from certain groups due to hardware, software, or skills constraints or deployment contexts (e.g. geographic region, internet speed, devices) & Better hiring and promotion pathways for people with access to generative AI models \citep{gmyrek_generative_2023}± \\ \hline Environmental damage                                       & Creating negative environmental impacts though model development and deployment & Increase in net carbon emissions from widespread model use \citep{patterson_carbon_2021}±                                                                                                        \\ \hline Inequality and precarity & Amplifying social and economic inequality, or precarious or low-quality work & Lower pay and precarious conditions for creative professionals (e.g. illustrators or sound designers) \citep{zhou_ai_2023}*                                                               
                                \\ \hline Undermine creative economies 
        & Substituting original works with synthetic ones, hindering human innovation and creativity
        & AI-generated artefacts leading to a homogenisation of aesthetic styles \citep{epstein_art_2023}± \\ \hline Exploitative data sourcing and enrichment          & Perpetuating exploitative labour practices to build AI systems (sourcing, user testing)  & Exposing human annotators to toxic audiovisual content \citep{perrigo_exclusive:_2023}*                                                              
\\ \bottomrule

\end{longtable}
\end{footnotesize}

References marked with (*) indicate real-world examples; those marked with (±) are hypothetical and indicate anticipated risks.

\clearpage

\twocolumn

\subsection{Evaluation methods per layer}
\label{appendixa2}

\subsubsection{Capabilities layer}
\label{appendixa21}

\paragraph\textbf{{Human annotation}}
\label{humanannotation}

While automated annotation is increasingly common, it is usually calibrated against \textit{human annotation}. Human annotation is taken as the ground truth for most risk areas\footnote{Exceptions include harms where reliable algorithmic measures exist, such as privacy harms which can be measured via “memorisation”, and debates are ongoing as to the quality of human annotation on factuality \citep{hosking_human_2023}. Human annotation is also often considered in conjunction with other, basic metrics such as equalised odds to measure fairness.} because whether a given AI system output is offensive, misleading, or indicates other risks of harm is subject to human judgement. Human annotators are shown a series of model outputs and tasked to judge these against a given set of criteria, such as a pre-existing set of rules or content policies (e.g. \citet{glaese_improving_2022, thoppilan_lamda:_2022}). For example, human annotators can leverage expertise to assess whether a given model output is likely to polarise political debates or sow division between social groups, although targeted expert annotation is relatively rare. Different annotators may provide different types of domain expertise, such as medical expertise or different lived experiences \citep{abercrombie_risk-graded_2022}. To assess social and ethical harms, human annotators can assess whether a given AI system output indicates risks of downstream harm – for example, by showing factually incorrect, offensive, or misleading content.

Human annotation presents a number of limitations. Depending on the annotation task, annotators may be exposed to risks of harm \citep{stoev_experiencing_2023}. Human annotators are often employed under precarious conditions \citep{gray_ghost_2019}. While annotation that leverages specific forms of human expertise is valuable, it can be difficult to source via commonly used annotation platforms \citep{zhang_needle_2023}. To AI developers, human annotation is comparably costly and time-intensive. 

While human annotation is often considered the gold standard, annotated data still has limits. These stem from unrepresentative annotator pools, incentives that drive down quality, and poor design of annotation interfaces or tasks. Demographic characteristics such as human raters’ self-described identities can significantly impact their ratings, raising concerns about introducing biases into annotated datasets (e.g. \citet{aroyo_reasonable_2023, homan_intersectionality_2023, goyal_is_2022}). This is especially problematic considering that the demographics of selected human annotators to fine-tune or evaluate generative AI models are rarely representative of the broader population (see, for example, \citet{bai_constitutional_2022, ouyang_training_2022}). To signpost such limitations, it is advisable to analyse and disclose data annotator demographics to indicate whose perspectives have been included in a dataset, as a basic proxy for what biases are likely to exist in that dataset \citep{aroyo_dices_2023, sap_annotators_2022, prabhakaran_releasing_2021}. 

Concerns have also emerged with respect to aggregating schemes commonly used to compile diverse labels from human annotators into a single ground truth label. These practices have been criticised for leading to an overestimation of future model performance \citep{gordon_disagreement_2021}, while bypassing important views or disagreements from minority raters on sensitive topics \citep{marchiori_manerba_bias_2022, field_survey_2021}. Gathering human annotations from online crowdsourcing platforms such as Amazon Mechanical Turk (MTurk), Scale, or Surge is common for large-scale annotation tasks. However, crowdsourcing can be associated with issues of annotation quality, as raters are financially incentivised to bypass certain tasks (e.g. skipping questions) or complete them as fast as possible \citep{organisciak_evaluating_2012}. Data quality deteriorates as human annotators themselves use AI systems to assist the data annotation work \citep{veselovsky_artificial_2023}. Possible ways to mitigate these concerns include equitable pay, payment by time not by task, attention checks, and annotation quality thresholds \citep{jindal_implementing_2022, abbey_attention_2017}. Careful design of annotation tasks leveraging expertise from psychology experiment set-ups can augment data quality. 

\paragraph\noindent\textbf{{Benchmarking}}
\label{benchmarking}

The most common evaluation method of model capabilities is benchmarking, which assesses AI system performance against a predefined task, such as mapping AI system outputs to a dataset of prompts and responses. Benchmarks are “success tests”, meaning that each task has a clear intended outcome against which model performance is measured. They differ from exploratory tests, where model patterns or “signatures” are the focus \citep{taylor_signature-testing_2022}. Benchmarks may explicitly target failure cases that were identified in a given AI system, thus constituting bespoke challenge datasets. In addition to assessing model output, benchmarks can also be used to assess technical performance metrics, such as energy use at inference (e.g. \citet{kaack_aligning_2022, wang_benchmarking_2020}). Benchmarks have been applied to indicate a range of ethical and social risks – for example, by measuring the likelihood that an AI system outputs toxic or discriminatory content \citep{cho_dall-eval:_2023, gehman_realtoxicityprompts:_2020}. Benchmarks may be generated from previously annotated data (see \hyperref[humanannotation]{Human annotation}) \citep{svikhnushina_approximating_2023}. 

Benchmarks ensure test-retest reliability\footnote{Test-retest reliability means that running the same benchmark twice should, in theory, lead to the same results. It is undermined where a model has been silently iterated on or fine-tuned in between measurements \citep{tu_chatlog:_2023, chen_how_2023}.} and while they incur computational and financial cost \citep{perlitz_efficient_2023, liang_holistic_2022}, they can be comparably cost-effective and time-effective for AI developing organisations,\footnote{Automated benchmarks require technological expertise to set up and incur computational cost, though these are often readily available to large organisations developing generative AI systems. Evaluating smaller versions of an AI system can provide some indication of potential capabilities and risks, but is insufficient as risks may also emerge or become more salient with model size \cite{wei_emergent_2022}.} making them suitable for frequent use. Regular testing of an AI system against benchmarks permits tracking model progress over the course of training and allows for cross-model comparisons where metrics are shared in the field. AI developers may run benchmarks regularly over the course of model development to obtain a frequent signal of AI system capabilities that guides AI design (“hill-climbing”). Benchmarks are also used to inform responsible decision-making. Sometimes, evaluations that were initially created for the purposes of responsible decision-making come to be used for “hill-climbing” to bake in ethical and sociotechnical considerations from the beginning; here it is important to keep track of which aim an evaluation is intended to perform and whether it is “held out” or may fall prey to Goodhart’s Law as discussed below (see \hyperref[rolesandresponsibilities]{Roles and responsibilities}) \citep{liao_rethinking_2023, bolukbasi_man_2016, gonen_lipstick_2019}.

With increasingly anthropomorphic design in generative AI systems and the development of synthetic relatable characters and simulacra \citep{griffith_my_2023, pentina_exploring_2023, park_generative_2023}, it is tempting to evaluate these systems through methods that were originally devised to assess human or animal psychology \citep{binz_using_2023, bubeck_sparks_2023, frank_baby_2023}. The transfer of cognitive tests to the evaluation of AI has a long history that precedes generative AI systems (e.g. \citet{crosby_animal-ai_2019, noauthor_bridging_nodate}). Such psychology-inspired experimentation and benchmarking has been applied to assess whether generative AI systems display certain cognitive capacities, such as Theory of Mind \citep{sap_neural_2022} or are “cooperative” \citep{chan_towards_2023}. However, it is questionable whether applying tests to study constructs such as “empathy” in humans yield any valid or meaningful results when applied to AI systems that are so fundamentally different from human minds \citep{shiffrin_probing_2023, ullman_large_2023}. Tests that were developed for studying animal cognition or the human mind rely on a range of assumptions (e.g. regarding life cycles, ballpark estimates of memory and learning capacities, and embodiment) which may not hold for AI systems \citep{narayanan_gpt-4_nodate, mitchell_how_2023}. Another problem is that established benchmarks including such experiments suffer validity problems due to “memorisation”, where the correct answers may have inadvertently been learned from textual descriptions in AI assistant training data \citep{schaeffer_are_2023, de_wynter_evaluation_2023, mitchell_how_2023}. 

Benchmarking as a method to evaluate AI systems faces a range of further limitations. First, benchmark datasets can face the same limitations as any other dataset and be too small, narrow, or biased for the task at hand. Like any other capability layer evaluation, benchmarks are limited in assessing ethical or social risk because they do not capture important context. Benchmarks are limited datasets and often permit narrower inferences than their title implies \citep{raji_ai_2021, schlangen_language_2019, narayanan_gpt-4_nodate}. This explains why benchmarks may yield incompatible results: one paper describes the “benchmark lottery”, whereby model performance may seem high on one benchmark and low on another benchmark purportedly testing the same construct \citep{dehghani_benchmark_2021}. To avoid narrow benchmarks and running into Goodhart’s Law (see \hyperref[rolesandresponsibilities]{Roles and responsibilities}), one approach is to create “dynamic”, continuously evolving benchmarks, where the underlying datasets are updated over time (see, for example, GEM, \citet{gehrmann_gem_2021, kiela_dynabench:_2021}). An additional, emerging practice uses pre-trained models to evaluate the outputs of other AI systems \citep{openai_gpt-4_2023-1, bai_constitutional_2022, wu_human_2023}. This enables dynamic testing and creates some degree of flexibility in what is evaluated. This approach could operationalise a user-specified definition of harm, supplied via a prompt or small fine-tuning dataset, thus creating a versatile avenue for highly bespoke probing of an AI system. This automatised approach allows for faster iteration on how a harm is operationalised and can be simpler than collecting an entirely new static evaluation dataset for each new definition or operationalisation of harm. However, this approach is limited by its use of the same type of models the evaluation seeks to measure. Without grounding in human evaluation, there is a risk of amplifying harmful behaviours that models exhibit but cannot identify (see \hyperref[modeldriven]{Model-driven evaluation may fill gaps}).

A second set of challenges is that benchmarks are often flawed or constructed without necessary care and documentation, raising validity concerns \citep{blodgett_stereotyping_2021, liao_are_2021}. To overcome the narrowness problem, multiple benchmarks can be aggregated into a “test suite” (e.g. HELM, \citet{liang_holistic_2022}, BIG Bench, \citet{srivastava_beyond_2023}, Safetykit, \citet{dinan_safetykit:_2022}). However, more complex benchmarks may still not capture relevant context or aspects of the target construct. Furthermore, collapsing multiple tests into a single result can make it harder to interpret results and can mask issues such as disparate performance for different groups, as scores are aggregated \citep{burnell_rethink_2023}.

Benchmarks are also limited in terms of external validity – that is, with respect to how well results generalise to novel instances \citep{de_vries_towards_2020}. This problem is particularly pronounced in benchmarks that transpose tests designed for human cognition to AI systems, without checking that underlying assumptions hold \citep{binz_using_2023, bubeck_sparks_2023, shiffrin_probing_2023, schaeffer_are_2023, de_wynter_evaluation_2023, mitchell_how_2023}. 

Human annotation can be used to develop automated benchmarks, which simulate human ratings on novel outputs. Overall, there is a push toward automating testing to reduce labour and time cost, and to reduce human exposure to potentially harmful model outputs. However, automated tests built on human annotation introduce additional noise, as they often fail to accurately represent human judgements and are inevitably less accurate \citep{van_miltenburg_gradations_2020, liu_how_2017, novikova_rankme:_2018}. This situation can then be further compounded if human annotators themselves use AI systems to assist the data annotation work \citep{veselovsky_artificial_2023}. Recent participatory efforts aim to address these limitations by creating avenues for meaningful input from people from different cultural backgrounds and better benchmarks to identify cultural tropes or stereotypes in images \citep{qadri_towards_2023}. 

Finally, benchmark testing alone may leave unknown capabilities and failure modes undetected. While AI developers have the most technical knowledge about how generative AI models work, there is still a lack of understanding of what these models are capable of (also referred to as “capability overhang”, \citet{shevlane_model_2023}). This makes it particularly difficult to foresee new model abilities – properties that materialise as the complexity of the system increases \citep{wei_emergent_2022} – that may cause downstream risk. These may not be possible to predict simply by extrapolating the capabilities of smaller-scale models \citep{wei_emergent_2022, schaeffer_are_2023}. 

\paragraph\noindent\textbf{{{Adversarial testing}}}
\label{adversarialtesting}

Adversarial testing, also referred to as “red teaming”, refers to risk identification exercises through adversarial attacks on AI models, infrastructure, development and deployment environments, and deployed AI products/systems. It is a mode of evaluation targeted at finding vulnerabilities in AI systems that can be exploited to get an AI system to output harmful content \citep{brundage_toward_2020, casper_explore_2023, ganguli_red_2022, wei_jailbroken:_2023, milliere_adversarial_2022}. These attacks can be conducted in any modality or across modalities. They are particularly suitable for assessing model vulnerabilities or the friction users encounter when seeking to obtain harmful material. For example, targeted attacks based on textual inputs can be used to produce harmful image outputs \citep{yu_adversarial_2023, ma_understanding_2021}. Adversarial testing can be conducted by humans or it can be automated, such as by leveraging large language models to perform adversarial testing \citep{perez_red_2022, yang_sneakyprompt:_2023, zou_universal_nodate}. Much like automated benchmark development based on human data, successful adversarial testing prompts can be reused to automate future testing.

Open-ended probing is a method for surfacing unexpected risks and risks that arise in response to innocuous outputs. In the context of safety evaluation, such exploratory probing is a form of “red teaming”: researchers may centre a given risk area, such as eating disorder content, and then perform exploratory prompting of AI systems to explore biases and patterns in model responses \citep{ccdh-eating-disorders}. Exploratory safety evaluation may also include the visualisation of thematic clusters in training data to help surface potential risks of representation harms (KnowYourData, \citet{google_know_nodate} REVISE, \citet{wang_revise:_2021}). Exploring the provenance of data can indicate unexpected risks – for example, to what extent private or copyrighted data is present in the dataset, data which may in turn be leaked or cause privacy risks \citep{dodge_documenting_2021, choi_tools_2023, wang_measuring_2022, kreutzer_quality_2022}. Open-ended probing is limited in that it does not assess specific harm areas or test hypotheses about particular AI system capabilities. It is often a helpful first step to identify leads for more directed evaluation.

Adversarial testing more broadly is subject to the same limitations as human testing and benchmarking outlined above. In addition, adversarial testing is limited by the imagination, contextual knowledge, and skill as probers seek to compromise the AI system. Despite adversarial testing, novel failure modes are typically discovered when an AI system is released to the broader public \citep{wei_jailbroken:_2023}. Automated red teaming presents a highly scalable approach but so far has yielded lower-quality tests than human annotation. Further assurances on the quality and reliability of such tests are therefore needed \citep{perez_discovering_2022, mozes_towards_2023}.

\subsubsection{Human interaction layer}
\label{appendixa22}

\paragraph\noindent\textbf{{{Behavioural experiments}}}
\label{behaviouralexperiments}

Pathways by which an AI system may cause harm to the person interacting with it can be assessed through controlled studies leveraging psychology, human–computer interaction, or behavioural economics methodologies \citep{lee_evaluating_2023, tahaei_human-centered_2023}. These methods isolate variables of interest, often in highly controlled settings. They may target outcomes from interactions or mechanisms by which potential harm may occur – such as the common propensity for people to believe misinformation more readily if it aligns with their views \citep{lodge_rationalizing_2013}. Behavioural experiments typically aim for some level of generality, rather than isolating specific use cases or application designs. This lends them to the study of potential impacts through human interaction of “general purpose” technologies, such as AI systems whose application domain and product design is not yet defined.

Experiments face a realism trade-off whereby highly contrived laboratory experiments give better insight into causal mechanisms but may not extend to real-world use cases. Another challenge is that such experiments are typically limited in the number of participants, and small-scale studies may fail to detect harms with small effect sizes, which may only become visible at scale – requiring either passive monitoring or other forms of larger-scale analysis. Results may be further confounded due to unrepresentative participant pools, which are often white, educated, and from industrialised, rich, and democratic locales \citep{henrich_weirdest_2010}. Similarly to human annotation, behavioural experiments have among their practical limitations the fact that they are time-, labour-, and cost-intensive, and may expose human participants to potential harm. Such experiments also require expertise in experiment design, user interfaces, and internal review processes suited to the ethical considerations that arise in this kind of research, some of which may be rare in organisations that develop AI systems \citep{jackman_evolving_2016, zevenbergen_internet_2020, jindal_implementing_2022}.

\paragraph\noindent\textbf{{{User research}}}
\label{userresearch}

User research and usability testing can evaluate the needs and behaviours of users, as well as the functionality and possible externalities that AI systems may create at the point of use.\footnote{Note that these tests are not often subject to the same ethics scrutiny as psychology experiments. This has led to user studies that adversely impacted user well-being \citep{jouhki_facebooks_2016}.} It can be conducted in contrived environments to isolate specific variables, such as the impact of particular features or the user interface of applications. User studies can also be conducted “in the wild”, by observing how people use AI systems in real-world scenarios. User testing may include behavioural experiments but can also include interviews, talk-out-loud studies, or surveys. 

Experimental user studies resemble psychology and human–computer interaction testing as they require human participants whose experiences are studied either in controlled laboratory conditions or in real-world contexts. Differently from behavioural experiments, user testing focuses on specific applications in particular contexts. As a result, the findings often do not extend to other AI systems. Experimental user testing methods include A/B tests to isolate potential impacts of an AI system design feature. These methods can be used to assess how people actually use an AI system, whether it works equally well for different user groups, and whether interaction with the AI system presents unexpected safety hazards. User testing can also measure risk trajectories over time – for example, by mapping out the risk of people being exposed to harmful content over the course of “user journeys”, i.e. over the course of human interaction with an AI system. Such user journeys can also model trajectories of users with malicious intent \citep{roy_determining_2023}. “Adversarial testing”, while primarily used to investigate performance failures of AI systems, can focus on particular use cases and shed further light on the friction that people may encounter when trying to use the AI system to malicious ends.

Passive monitoring in user studies may include observing patterns in participant interaction with AI systems, such as when, why, and how people try to put the AI system to use. Such passive monitoring can usefully be complemented by qualitative methods, such as interviews or encouraging participants to “talk out loud” as they use a given application. These, in turn, can be complemented by quantitative approaches, such as surveys. Such mixed-methods approaches can triangulate multiple sources of data for a more nuanced understanding of how people use AI systems and what potential unintended externalities they may experience.

User testing has historically focused on product development, rather than risk evaluation for assurances and to inform responsible decision-making. Or, where such user testing may occur, it is not commonly publicly disclosed, and there is thus lacking a precedent to compare new AI systems against. However, user testing is increasingly necessary for safety evaluations, given the growing concern about potential ethical and social risks from human interaction with generative AI systems – such as overtrust and overreliance; anthropomorphism risks and potential emotional harm; and well-evidenced problems of disparate functionality and externalities between different user groups. AI system and product developing organisations often host user testing expertise and infrastructure. These groups could expand their remit to perform human–AI interaction evaluations to help assess potential risks of harm. As the limitations of capability testing for risk evaluation are becoming clear, more evaluations at this layer may be conducted in the future.

\paragraph\noindent\textbf{{{Passive monitoring of human use}}}
\label{passivemonitoring}

The passive monitoring of people’s activity on the platform where the AI system is deployed can indicate ethical or social harm. Passive monitoring can be done during pilot releases or after wide-scale deployment. Collected data may reveal effects during an interaction, as well as effects that last beyond the human–AI interaction. For example, prior research on human–AI teaming used a platform where humans could interact with an AI teacher to improve the skill of playing Go. This work identified effects of the human–AI interaction on human skill and playing style that showed after the human had interacted with an AI system \citep{choi_how_2022, shin_human_2021}. Passive monitoring also includes mapping out risk over the course of a “user journey”, such as the risk of being exposed to misinformation when seeking information on a search engine \citep{roy_determining_2023}.

For such studies, quasi-experimental data may permit inferences on causal relationships without conducting active interventions. Such work can leverage expert annotation on how human experts interact with AI systems in particular domains or tasks. There are also beginning to be automated tools for evaluating mechanisms in human–AI collaboration, such as an automated approach to detecting patterns of turn-taking in human–AI collaborative writing \citep{zeng_towards_2023}. Passive monitoring can be combined with active interventions for more robust results. For example, one study triangulated reviews from an app store with a follow-up survey to assess problems in user–AI interaction \citep{eiband_when_2019}. Passive monitoring also lends itself to longitudinal studies, such as assessing transcripts and interaction data to observe human emotional attachment to AI companions and, more broadly, human relationship-building with generative AI systems \citep{xie_attachment_2022, pentina_exploring_2023}.

The main limitation of passive monitoring is that it rarely allows the isolation of causal effects, as it makes no direct experimental intervention. To evaluate potential harmful impacts, further experimental testing under controlled conditions may be necessary. 

\subsubsection{Systemic impact layer}
\label{appendixa23}

\paragraph\noindent\textbf{{{Staged release and pilot studies}}}
\label{stagedrelease}

Staged release processes or pilot studies can give insight into potential systemic impacts. An AI system can be deployed in a controlled setting. For example, a business, hospital, or public institution may adopt an AI system as a pilot test, with close monitoring to assess impacts such as on the provision of care \citep{elish_repairing_2020}. Such monitoring may use ethnographic methods that entail witnessing how a group of people use an AI system in a real-world setting alongside qualitative methods such as interviews. Ethnographic methods can reveal systemic impacts of an AI system as it is adopted in institutions. For example, on AI systems other than generative AI, ethnographic methods revealed how embedding an AI system amplified unfair discrimination in a specific police force \citep{marda_data_2020}. Ethnographic methods can also shed light on the functionality of an AI system at the point of use, i.e. an evaluation at the human interaction layer. 

Pilot release studies may also constitute a safe environment for running experiments to isolate causal effects. For example, experimental roll-outs of AI systems have tested the impact on productivity of an AI system in the workplace \citep{brynjolfsson_generative_2023}. Human–AI interaction experiments as described above can also constitute a form of pilot study that targets systemic impact. This has been done to assess potential economic impacts, such as productivity impacts of co-writing with ChatGPT \citep{noy_experimental_2023} or developer productivity and happiness when co-programming with an AI \citep{peng_impact_2023}. Other experimental approaches such as randomised controlled trials (RCTs) could also be applied to generative AI systems to identify impacts by comparing outcomes to organisations where AI systems are deployed differently or not at all. 

Staged release and pilot studies have two main weaknesses. First, similarly to experimental methods, as described above, staged roll-outs or pilot projects may not allow for generalisations to novel contexts. This is particularly the case for highly specific studies on local contexts, such as ethnographic work. Small-scale experiments may also yield misleading results – for example, they may identify impacts that are negated by equilibrium effects that emerge through large-scale adoption \citep{lise_equilibrium_2004}. Further, pilot studies that have ecological validity regarding intended deployment contexts require some level of deployment of an AI system, which raises potential safety concerns. Ethnographic studies also require embedding researchers in potential deployment contexts, which can create disruption or discomfort for local institutions or systems.

\paragraph\noindent\textbf{{{Impact assessments}}}
\label{impactassessments}

Impact assessments may be conducted before an AI system is deployed (e.g. algorithmic impact assessments (AIAs),  \citet{noauthor_algorithmic_nodate-1, noauthor_algorithmic_nodate}), or retrospectively (also referred to as ex-post assessments). Impact assessments can track potential effects on broader economic, environmental, or social structures. Prospective impact assessments may be grounded in guiding questions to assess potential risks of harm. Retrospective impact assessments can be grounded in the monitoring of different system-level indicators and patterns. For example, observational data may monitor different economic indicators to quantify the impact on employment and inequality of AI-adjacent technologies such as robotics \citep{acemoglu_ai_2020, bonfiglioli_robot_2020, acemoglu_robots_2020}. Retrospective impact assessments may also leverage expert views, for example, by surveying or interviewing groups that may witness systemic impacts, as has been done for other types of AI \citep{hunt_measuring_2022, bell_ai_2023}. Interviewing different groups can further shed light on the distribution of systemic impacts from AI systems, such as one study finding increased negative emotions for low-skilled employees in contrast to increased creativity for high-skilled employees (e.g. \citet{jia_when_2023}). Instances of observed harm can be aggregated to identify patterns and systemic impacts \citep{noauthor_welcome_nodate}. Impact assessments also comprise case studies, for example, evaluating the impact of generative AI systems on the creative commons, a public good \citep{del_rio-chanona_are_2023, huang_generative_nodate}, or impact on job automation \citep{milanez_impact_2023}.

Impact assessments, differently from experiments and pilots, rely on qualitative and/or observational data about complex and prolonged processes, which makes it difficult to generate causal evidence of impact. This can be partially addressed by leveraging natural experiments such as policy shocks that introduce exogenous variation on technology adoption. However, this restricts the range of potential studies that can be undertaken and requires strong assumptions about the link between exogenous shock, technology adoption, and impact \citep{kiviet_testing_2020}. Perhaps even more importantly, the evidence from ex-post assessment arrives too late to guide model development and deployment to prevent real-world harms or avoid potential lock-in to harmful technologies \citep{acemoglu_regulating_2023}. Forecasts and experiments that can generate evidence of (potential) impact further upstream can help address this gap.

\paragraph\noindent\textbf{{{Forecasts and simulations}}}
\label{forecasts}

Forecasts and simulations of downstream impact can help evaluate risks of harm from AI systems on broader structures. For example, anticipated AI system capabilities have been mapped against tasks required to perform certain jobs, to forecast likely downstream impacts on labour markets \citep{autor_new_2022, webb_impact_2019, frank_toward_2019}. Comparative technologies can be evaluated as a proxy for potential impact from generative AI systems. Assessing their impact can help estimate impacts from a given AI system or serve as a starting point for determining what harms may arise and what mistakes should not be repeated \citep{autor_new_2022, webb_impact_2019}. Where no highly similar technologies exist, loose analogies may be useful heuristics\footnote{For example, AI systems have been likened to other technological innovations such as steam engines and office automation \citep{acemoglu_power_2023}.} but should be used with caution. 

Forecasts rely on a range of ambiguous decisions, which lead to uncertain results. For example, in evaluations of economic exposure to automation based on analysis of the overlap between model capabilities and work tasks, much depends on how model capabilities are defined and quantified, making transparency in data sources and methods critical \citep{tolan_measuring_2021, frank_toward_2019, frey_future_2017, felten_occupational_2021, eloundou_gpts_2023}. By way of illustration, in the case of labour market forecasts, it is generally difficult to estimate if exposure will translate into a positive impact (e.g. a worker becomes more productive) or a negative one (a worker is displaced) without additional assumptions about the deployment context or ex-post analyses.

\subsection{Case study: Misinformation}
\label{casestudy1}

Misinformation is defined as the spread of false, inaccurate, or misleading information, often unintentionally \citep{wardle_information_nodate}.\footnote{This is distinct from disinformation, which is false information created with the explicit intent to deceive or cause harm to others \citep{wardle_information_nodate}.} Large multimodal AI systems can produce highly realistic but factually incorrect or misleading content, such as images of events that never took place or synthetic audiovisual representations of people saying things they never said (\citep{birnbaum_ai_2023, weise_when_2023}. These are often described as “confabulations” or “hallucinations” when generated in response to an innocuous query \citep{ji_survey_2023, li_evaluating_2023, xiao_hallucination_2021}. False or misleading information can also be created and spread with the explicit intent to deceive – as part of large-scale disinformation campaigns, for example \citep{satariano_people_2023, marwick_media_2017}. 

Factually inaccurate or fictitious outputs may be harmless, such as when used for creative or satirical purposes \citep{kasirzadeh_conversation_2023, diaz_accounting_2022}. However, in other contexts, misinformation generated from multimodal AI systems may pose a significant risk of harm, such as leading people to act on false beliefs, including in legal, medical, or other high-stakes contexts. People often struggle to reliably distinguish synthetic from human-generated content \citep{spitale_ai_2023, nightingale_ai-synthesized_2022, kreps_all_2022, groh_deepfake_2022, clark_all_2021}. Audiovisual misinformation spreads faster online and is perceived as more credible than text-based misinformation \citep{hameleers_picture_2020, sundar_seeing_2021}, exacerbating potential downstream risks. In May 2023, for example, an AI-generated image that was widely circulated on Twitter led several news outlets to falsely report that an explosion had taken place at the US Pentagon, causing a brief drop in the US stock market \citep{alba_how_2023}. 

AI-generated misinformation could also have broader societal repercussions, such as eroding public trust in evidence and information. As synthetic audiovisual content becomes more widespread, research suggests that people may become more uncertain about what to believe and, as a result, more distrustful of established sources of information \citep{vaccari_deepfakes_2020, lee_antecedents_2023}. A related concern is that inaccurate or misleading synthetic outputs created with generative AI systems could end up contaminating open “knowledge commons” (e.g. Wikipedia), jeopardising the quality and legitimacy of shared public knowledge \citep{huang_generative_nodate}. This could also undermine the authentication of evidence by making it easier for people to reject genuine audiovisual evidence of wrongdoing as “fake” \citep{bond_people_2023, citron_deep_2019, pfefferkorn_deepfakes_2020, pawelec_deepfakes_2022}. 

Evaluating \textit{misinformation} harms comprehensively requires measuring different concepts at each of the three layers of a sociotechnical approach to evaluation. To make multifaceted harms such as misinformation tractable, it can be helpful first to identify concepts that relate to or constitute the overall harm. Such concepts sit between the latent construct (e.g. “misinformation”) and the concrete measure (e.g. FID scores). An example is “factuality”. Factuality is conceptually narrower than the high-level concepts of harm, but it is broader than the specific metrics that may be used to measure it (e.g. FID scores). Adding this intermediate step of identifying narrower concepts that compose a harm can help build the bridge between latent constructs and concrete metrics. It can also indicate where an evaluation may be able to shed light on multiple risk areas.

\subsubsection{Capability}

\paragraph{Operationalisation: Factuality of AI system outputs.} The likelihood of an AI system generating misinformation can be assessed via benchmarks and other tests at the capability layer. Different methods have been proposed to assess and improve the factual accuracy of outputs from generative AI systems. Due to the widespread availability of textual data, most focus on the text modality; in contrast, evaluations of factuality for image, audio, and video modalities are sparse. 

Several approaches involve automatic \textit{fact verification} – i.e. fact-checking AI system outputs against existing knowledge sources (e.g. Wikipedia articles or databases of factual statements such as WikiData) \citep{openai_gpt-4_2023-1, lee_factuality_2023}. Others focus on \textit{verifiability}, testing whether statements generated by an AI system can be attributed to reliable external sources (FActScore, \citet{min_factscore:_2023}; RARR, \citet{gao_rarr:_2023}; \citet{liu_evaluating_2023}). Wiki-FACTOR and News-FACTOR \citep{muhlgay_generating_2023} as well as FactualityPrompt \citep{lee_factuality_2023} test an AI system’s factual knowledge, i.e. its propensity to generate factual statements from a corpus versus incorrect ones, while TruthfulQA measures the likelihood of its generating false answers learned from imitating human texts in response to open-ended questions \citep{lin_truthfulqa:_2022}. Recent work in this area focuses on improving factuality by training AI systems to automatically detect false claims, multimodal misinformation, and conspiracy theories (\citet{papadopoulos_synthetic_2023}; HaluEval, \citet{li_halueval:_2023}; HaDes, \citet{liu_token-level_2022, russo_acti_2023}).

However, these benchmarks capture a limited operationalisation of “truthfulness”. In practice, a more nuanced and contextual evaluation of factuality often requires human expertise. Some evaluations involve testing the AI system’s propensity to generate false and potentially harmful narratives \citep{brewster_could_nodate, noauthor_googles_nodate}. Most human annotation focuses on evaluating whether AI system outputs can be supported by references. There are two main approaches to this human annotation. One is to instruct annotators to fact-check the output against existing knowledge bases – by performing a Google or Wikipedia search, for example \citep{maynez_faithfulness_2020} – or to rate the degree of confabulation within a range \citep{ji_survey_2023}. The other approach is to measure the proportion of AI system responses that can be attributed to identified sources (AIS) \citep{rashkin_measuring_2022}. However, these approaches do not account for source quality and trustworthiness. 

Identifying and evaluating outputs for truthfulness present new challenges for audiovisual modalities, as this type of content may not be as easily verifiable against existing public knowledge bases. Some knowledge bases exist, such as Fakeddit \citep{nakamura_fakeddit:_2020}, Factify \citep{mishra_factify:_2022}, and Fake2M \citep{lu_seeing_2023}. Several datasets also exist to aid the detection of audio and video deepfakes (\citet{noauthor_datasets_nodate}; FakeAVCeleb, \citet{khalid_fakeavceleb:_2022}). Even so, the lack of large, up-to-date datasets for multimodal verification remains a major roadblock. For audio and image generation, as well as multimodal summarisation, alignment metrics e.g. CLIPScores \citep{hessel_clipscore:_2022}, CLAPScores \citep{radford_learning_2021, elizalde_clap:_2022}, and CLIPBERTScore \citep{wan_evaluating_2022} can help evaluate how well an AI system represents the relationship between audio or images and text \citep{otani_toward_2023}, though these metrics are quite general and not suitable for the detection of AI-generated audiovisual content. 

Manually checking the factual accuracy of AI system outputs is a laborious task that hinges on raters’ prior knowledge and fact-checking abilities, and on the availability of observable evidence against which to verify information. This is all the more complicated in cases where notions of “truth” are debated (such as political matters) or on topics where consensual knowledge is emerging and subject to change (e.g. Covid-19) \citep{kasirzadeh_conversation_2023, evans_truthful_2021}. For that reason, factuality evaluations can quickly become outdated. While it is possible to train an AI system to not output content that contradicts existing scientific facts – assertions of flat earth theory, for instance – this is more challenging for novel content that is not already present in some existing data. 

\paragraph{Operationalisation: Credibility of AI system outputs.} Across all modalities, one key challenge of existing approaches to evaluating misinformation harms is that content factuality is not a sufficient proxy for determining whether a piece of content has the potential to misinform.\footnote{A common form of audiovisual misinformation, for example, is decontextualisation – that is, using real images to provide evidence in support of a false claim \citep{weikmann_visual_2022}.} Some fictitious or factually incorrect content may be harmless because it defies common sense and is therefore unlikely to be believed at face value (e.g. a video of a cat scuba diving). Conversely, some types of outputs have a higher chance of deceiving or misinforming audiences if they are perceived as credible or believable, even if they are not factually accurate \citep{klein_plausibility_2023}. One way to evaluate the credibility of multimodal outputs at the AI system layer is to measure their perceptual quality or realism. Automated tools assess quality and fidelity scores (e.g. FID scores, Inception scores), which measure how closely generated images and audio resemble real reference outputs \citep{heusel_gans_2018,Salimans_Goodfellow_Zaremba_Cheung_Radford_Chen_2016}. However, fidelity scores in particular have been criticised for correlating poorly with human perceptions of quality \citep{saharia_photorealistic_2022}. 

Human evaluations also tend to focus on perceptions of image and speech quality (e.g. Mean Opinion Scores of the realism and perceptual quality of synthetic outputs compared to natural images) \citep{otani_toward_2023, kong_diffwave:_2021}. Other benchmarks specifically test human ability to distinguish real and AI-generated images – an example of a benchmark that targets usability layer outcomes (HPBench, \citet{lu_seeing_2023}). Common datasets and benchmarks used for qualitative benchmarking of image generation AI systems include Drawbench, MSCOCO and PartiPrompts \citep{saharia_photorealistic_2022, yu_scaling_2022, lin_microsoft_2014}. Perceptions of realism and credibility can be influenced by a number of factors, however, including an individual’s prior beliefs, knowledge, and trust in the source of the message, and the context in which they see the message. As a result, the credibility of a particular AI system output and how likely it is to deceive end users will differ based on user demographics and AI use cases. Disentangling these mechanisms requires targeted user testing at the human–AI interaction layer.

\paragraph{Operationalisation: Societal relevance of AI system outputs.} Compelling AI-generated synthetic content may run a greater risk of causing harm in politically salient contexts or during particular periods, such as prior to an election. AI-generated videos of police brutality or protests, for example, have the potential to sow division and shape public opinion about political events \citep{taylor_amnesty_2023}. Developing evaluation metrics for “social relevance” or “newsworthiness” of AI system outputs at this layer is therefore an important step toward evaluating potential downstream harms. Operationalising these terms into concrete metrics is often difficult in practice, as these are complex concepts whose interpretation differs based on locale. Expert fact-checkers, journalists, and misinformation experts have the necessary expertise to establish a set of criteria to guide the evaluation of multimodal AI system outputs that pose a high misinformation risk.\footnote{Similar efforts have been undertaken for automatic claim detection \citep{konstantinovskiy_toward_2021} and automatic detection of potentially malicious threads on the DarkWeb \citep{jin_darkbert:_2023}.} These efforts could consider factors like the domain of the content generated (e.g. legal, medical, or news-related) or the likelihood that a specific piece of content will be used to support factual statements or claims. Expert evaluation might also assess the emotional salience of AI system outputs (e.g. whether it evokes fear, anger, or moral outrage), as this is widely considered a key characteristic of persuasive misinformation \citep{crockett_moral_2017, han_anger_2020}. However, the gold standard to assess this would be to study the behavioural reactions of people exposed to these types of content.

\subsubsection{Human interaction}

\paragraph{Operationalisation: Deception.} A key concept to measure at this layer is whether AI system outputs actually deceive end users \citep{park_ai_2023}. Recent efforts in this area focus on testing people’s ability to distinguish different types of outputs as synthetic or human-generated and their ability to identify whether they contain misinformation. Several studies find that people are more easily deceived by AI-generated than human-generated misinformation \citep{spitale_ai_2023, nightingale_ai-synthesized_2022, kreps_all_2022, groh_deepfake_2022, clark_all_2021}. Others investigate the deceptive qualities of AI-generated content. Using qualitative and linguistic analysis, \citet{zhou_synthetic_2023}, for example, find that AI-generated and human-generated misinformation present significant linguistic differences, and that AI-generated misinformation is adept at “mimicking the attributes of existing information assessment guidelines”, including credibility and comprehensiveness. Generative AI systems that mimic speech patterns of users are also more often believed \citep{chiesurin_dangers_2023}. Finally, several studies investigate cues of deceptive behaviour, such as facial expressions and gestures, in multimodal audiovisual content (e.g. \citet{soldner_box_2019, zhang20c_interspeech}).

\paragraph{Operationalisation: Persuasion.} Other evaluations at this layer might focus on testing whether interacting with an AI system influences a person’s beliefs, attitudes, and behaviour \citep{deverna_artificial_2023}. This can be tested by measuring whether exposure to synthetic outputs convinces people to change their political views, or leads them to develop false or inaccurate beliefs about external reality \citep{dehnert_persuasion_2022}. For example, \citet{bai_artificial_2023} find AI-generated texts to be more persuasive than human-generated ones in shifting people’s attitudes on a range of policy issues. \citet{goldstein_can_2023} test the persuasiveness of AI-generated political propaganda through a series of survey experiments. 

Human-centred evaluations and experiments could also shine light on important underlying mechanisms and contextual drivers of misinformation, such as the extent to which different users are persuaded by different types of synthetic outputs. Prior studies identify several cognitive, social, and demographic factors that influence people’s receptivity to misinformation and propensity to develop false beliefs – factors such as partisanship, cognitive biases, or trust in information sources (for a review, see \citet{ecker_psychological_2022}). For example \citet{pennycook_prior_2018} demonstrate that repeated exposure increases people’s belief in misinformation. \citet{lovato_diverse_2023} find that people are better at identifying deepfake videos where synthetic personas match their demographics. Likewise, psychological studies show that emotional appeals increase the persuasiveness of misinformation and its impact on misperceptions \citep{martel_reliance_2020, tannenbaum_appealing_2015, lee_emotional_2023}.

Misinformation has the potential to deceive individuals not only by instilling falsehoods but by leading them to disbelieve true evidence and information \citep{rini_weaponized_2021}. It is therefore critical to test how interacting with an AI system impacts users’ belief in established facts and resistance to factual evidence. This is particularly important as mitigation techniques such as labels and watermarks to establish the provenance of AI-generated content become more widespread. Evidence suggests that labelling content as synthetic or “fake” could have unwanted consequences, such as increasing the perceived accuracy and believability of unlabelled content (otherwise known as the “implied truth effect”) \citep{pennycook_implied_2020}, and that these effects vary based on partisanship and label origin \citep{jia_understanding_2022}. These findings should be rigorously tested in the context of generative AI systems.

\subsubsection{Systemic impacts}
\paragraph{Operationalisation: Public trust in media and information.} Societal harms can be anticipated in foresight exercises and mitigated before they occur, but they can only be empirically measured at the third layer, once the AI system is deployed in society. This is because they require a certain amount of take-up and time before the effects manifest. To evaluate whether AI system adoption contributes to a broader erosion of trust at scale, population-level analysis of shifts in public trust can be conducted. Similar studies have been done in the context of social media and trust in mainstream information sources. \citet{park_global_2020}, for example, explore the impact of social media use on trust in news media globally. \citet{lorenz-spreen_systematic_2023} find strong correlational and causal evidence of a relationship between digital media use and low political trust, i.e. trust in media and political institutions. 

\paragraph{Operationalisation: Prevalence of synthetic content.} One way to test whether AI system adoption threatens to “pollute” and reduce the quality of publicly available information would be to measure the prevalence and spread of false or misleading synthetic content in the public domain post deployment. \citet{allen_evaluating_2020}, for example, evaluate the prevalence of fake news at the scale of the US information ecosystem, using nationally representative samples of mobile, desktop, and television-based media consumption – an approach that could be usefully extended to synthetic data. To test the impact of deploying a generative AI system on human-generated data on the open web, \citet{del_rio-chanona_are_2023} compared activity on Stack Overflow in locales with different levels of access to the AI system.

\paragraph{Operationalisation: Prevalence of synthetic content.} Another way to evaluate the likelihood of synthetic content contaminating the public domain at this layer is to test the effectiveness of methods for identifying AI-generated content, such as watermarking. Text-based watermarking is accomplished by adding hidden signals, such as tagging a subset of words, to the generated text \citep{kirchenbauer_watermark_2023}. For images and video, watermarks work by adding an imperceptible perturbation to an image’s pixel, creating an identifiable marker.\footnote{There have also been growing efforts around authentication of human-generated audiovisual content. See, for example, the work led by the Coalition for Content Provenance and Authenticity (C2PA).} 

AI systems used to create synthetic visual data embed unique traces in the output image, which can also be used to aid detection of fake images and attribute them to their different source AI systems \citep{sha_-fake:_2023}. AI labs – including Open AI, Google DeepMind, and HuggingFace, and public agencies like DARPA (e.g. through its SemaFor programme) – are currently developing automatic tools to help developers, key stakeholders, and the public better detect synthetic media. However, existing generative AI detection tools often struggle with generalisation, perform poorly on languages other than English, and are ill-equipped to deal with synthetic media that have been compressed or resized, as is common on social media \citep{liang_gpt_2023, corvi_detection_2022}. Watermarks are vulnerable to simple manipulation like resizing, cropping, or simply changing content format. These techniques therefore have to be continuously updated to prevent misuse.

\clearpage

\onecolumn

\addcontentsline{toc}{section}{Bibliography}
\bibliography{citations}

\end{document}